\documentclass[10pt,journal,compsoc]{IEEEtran}

\usepackage{graphicx}
\usepackage{amsmath}
\usepackage{amssymb}
\usepackage{booktabs}
\usepackage{multirow}
\usepackage{makecell} 
\usepackage{colortbl}
\usepackage{threeparttable}
\usepackage{pifont}
\newcommand{\cmark}{\ding{51}}

\usepackage[misc]{ifsym}

\newcommand{\etal}{\textit{et al}. }

\ifCLASSOPTIONcompsoc
  \usepackage[nocompress]{cite}
\else
  \usepackage{cite}
\fi

\ifCLASSOPTIONcompsoc
 \usepackage[caption=false,font=footnotesize,labelfont=sf,textfont=sf]{subfig}
\else
 \usepackage[caption=false,font=footnotesize]{subfig}
\fi

\usepackage{stfloats}

\usepackage[capitalize]{cleveref}
\crefname{section}{Sec.}{Secs.}
\Crefname{section}{Section}{Sections}
\Crefname{table}{Table}{Tables}
\crefname{table}{Tab.}{Tabs.}

\usepackage{multirow, overpic, textpos, array}
\usepackage{makecell}
\usepackage{footnote}

\usepackage{wrapfig,lipsum,booktabs}
\newlength\savewidth

\newcolumntype{x}[1]{>{\centering\arraybackslash}p{#1pt}}
\newcolumntype{y}[1]{>{\raggedright\arraybackslash}p{#1pt}}
\newcolumntype{z}[1]{>{\raggedleft\arraybackslash}p{#1pt}}

\usepackage{xcolor, colortbl}
\definecolor{Gray}{gray}{0.9}
\definecolor{baselinecolor}{gray}{0.9}

\newcolumntype{a}{>{\columncolor{Gray}}c}

\newcommand{\TableShotSOTA}{
\begin{table}[!t]
    \centering
    \setlength{\tabcolsep}{9pt}
    \renewcommand{\arraystretch}{1.1}
    \caption{Comparison with the shot-level state-of-the-art methods in terms of mAP(\%) on SoccerNet-v2. The second best for each column is underlined. }
    {
    \begin{tabular}{ccccc}
\toprule
\multirow{2}{*}{Method} & \multirow{1}{*}{Bound.}&\multicolumn{3}{c}{Transition} \\
    \cmidrule(l{1em}r{1em}){3-5}
  \iftrue
  & Det. & Abrupt & Fading & Logo \\
\midrule
  CALF~\cite{DBLP:conf/cvpr/CioppaDGGDGM20}    & 59.6 & 59.0 & \underline{58.0} & 61.8 \\
  Intensity~\cite{scikit}           & 64.0 & 74.3 & 57.2 & 28.5 \\
  Content~\cite{PyScenedetect}        & 62.2 & 68.2 & 49.7 & 35.5 \\
  Histogram~\cite{scikit}           & \underline{78.5} & \textbf{83.2} & 54.1 & \textbf{82.2}\\ 
  \midrule
  Temporal Perceiver & \textbf{81.9} & \underline{82.4} & \textbf{83.0} & \underline{78.8}\\
  \bottomrule
  \end{tabular}}
    \label{tab:camera-shots-results}
 \end{table}
}

\newcommand{\TableEventSOTA}{
\begin{table*}[ht]
\begin{minipage}{\linewidth}
\caption{Comparison with the event-level state-of-the-art methods in terms of f1@Rel.Dis on Kinetics-GEBD validation set and TAPOS validation set.}
\vspace{-2em}

\scalebox{0.95}{\centering
\setlength{\tabcolsep}{6.5pt}
\renewcommand{\arraystretch}{1.1}
\begin{tabular}{cccccccccccccc}&\\ \toprule
 \multirow{2}{*}{Benchmark} & \multirow{2}{*}{Method} & \multirow{2}{*}{Backbone} &\multicolumn{11}{c}{f1 @ Rel.Dis.}\\
 \cmidrule(l{0.5em}r){4-14}
  &   &   & 0.05 & 0.1 & 0.15 & 0.2 & 0.25 & 0.3 & 0.35 & 0.4 & 0.45 & \multicolumn{1}{c}{0.5}            &   avg          \\ \midrule
\multicolumn{1}{c}{\multirow{6}{*}{Kinetics-GEBD}}   & \multicolumn{1}{c}{BMN~\cite{DBLP:conf/iccv/LinLLDW19}}   & ResNet50         & 18.6        & 20.4 & 21.3 & 22.0 & 22.6 & 23.0 & 23.3 & 23.7 & 23.9 & \multicolumn{1}{c}{24.1}          &  22.3   \\
\multicolumn{1}{c}{}   & \multicolumn{1}{c}{BMN-StartEnd~\cite{DBLP:conf/iccv/LinLLDW19}} & ResNet50 & 49.1 & 58.9 & 62.7 & 64.8 & 66.0 & 66.8 & 67.4 & 67.8 & 68.1 & \multicolumn{1}{c}{68.3} &  64.0  \\
\multicolumn{1}{c}{} & \multicolumn{1}{c}{TCN-TAPOS~\cite{DBLP:conf/eccv/LeaRVH16, DBLP:conf/eccv/LinZSWY18}} & ResNet50 & 46.4  & 56.0  & 60.2  & 62.8  & 64.5  & 65.9  & 66.9  & 67.6  & 68.2  & \multicolumn{1}{c}{68.7} &   62.7
\\
\multicolumn{1}{c}{} & \multicolumn{1}{c}{TCN~\cite{DBLP:conf/eccv/LeaRVH16, DBLP:conf/eccv/LinZSWY18}} & ResNet50 & 58.8 & 65.7 & 67.9 & 69.1 & 69.8 & 70.3 & 70.6 & 70.8 & 71.0 & \multicolumn{1}{c}{71.2} &   68.5
\\
\multicolumn{1}{c}{} & \multicolumn{1}{c}{PC~\cite{DBLP:conf/iccv/ShouLWGF21}} & ResNet50 & 62.5  & 75.8  & 80.4  & 82.9  & 84.4  & 85.3 & 85.9  & 86.4 & 86.7  & \multicolumn{1}{c}{87.0} & 81.7 \\
 &  \multicolumn{1}{c}{{\color{gray}CVPR'22 LOVEU winner~\cite{DBLP:journals/corr/abs-2206-12634}\footnotemark[3]}} & {\color{gray}CSN} &{\color{gray}85.4} &{\color{gray}-} & {\color{gray}-}  & {\color{gray}-} & {\color{gray}-}&{\color{gray}-}&{\color{gray}-}&{\color{gray}-}&{\color{gray}-} & {\color{gray}-}&{\color{gray}-}
\\
\cmidrule(l{1em}r){2-14}
\multicolumn{1}{c}{}                          & \multicolumn{1}{c}{Temporal Perceiver}    & ResNet50     & \textbf{74.8}  & \textbf{82.8}  & \textbf{85.2}  & \textbf{86.6}  & \textbf{87.4}  & \textbf{87.9}  & \textbf{88.3}  & \textbf{88.7}  & \textbf{89.0}  & \multicolumn{1}{c}{\textbf{89.2}}          &   \textbf{86.0} 
\\ \multicolumn{1}{c}{}                          & \multicolumn{1}{c}{Temporal Perceiver}    & CSN     & \textbf{82.2}  & \textbf{88.0}  & \textbf{89.9}  & \textbf{90.9}  & \textbf{91.6}  & \textbf{92.0}  & \textbf{92.3}  & \textbf{92.5}  & \textbf{92.7}  & \multicolumn{1}{c}{\textbf{92.9}}          &   \textbf{90.5}
\\ \midrule

\multicolumn{1}{c}{\multirow{4}{*}{TAPOS}}  & 
\multicolumn{1}{c}{TCN~\cite{DBLP:conf/eccv/LeaRVH16, DBLP:conf/eccv/LinZSWY18}}    & ResNet50         & 23.7          & 31.2          & 33.1          & 33.9          & 34.2          & 34.4          & 34.7          & 34.8          & 34.8          & \multicolumn{1}{c}{34.8}          & 33.0                  \\
\multicolumn{1}{c}{}                          & \multicolumn{1}{c}{TransParser~\cite{DBLP:conf/cvpr/ShaoZDL20}}  & BNInception   & 28.9          & 38.1          & 43.5          & 47.5          & 50.0       & 51.4          & 52.7          & 53.4          & 54.0          & \multicolumn{1}{c}{54.5}          & 47.4          \\
\multicolumn{1}{c}{}                          & \multicolumn{1}{c}{PC~\cite{DBLP:conf/iccv/ShouLWGF21}} & ResNet50    & {52.2}          & {59.5}          & {62.8}          & {64.6}          & {65.9}       & {66.5}          & {67.1}          & {67.6}          & {67.9}          & \multicolumn{1}{c}{{68.3}}          & {64.2}          \\
\cmidrule(l{1em}r){2-14}
\multicolumn{1}{c}{}                          & \multicolumn{1}{c}{Temporal Perceiver}  & ResNet50   & \textbf{55.2}          & \textbf{66.3}          & \textbf{71.3}          & \textbf{73.8}          & \textbf{75.7}       & \textbf{76.5}          & \textbf{77.4}          & \textbf{77.9}          & \textbf{78.4}          & \multicolumn{1}{c}{\textbf{78.8}}          & \textbf{73.2}   \\
\bottomrule
\end{tabular}}
\label{table:compare_GEBD}
\begin{tablenotes}
\item \textit{The results are reported in percentage~(\%). The best results are highlighted in {\bf bold}. Temporal Perceiver achieves non-trivial improvement on average-f1 score as well as f1@0.05 on both benchmarks.}
\item \textit{\footnotemark[3] is the winner solution of CVPR'22 Generic Event Boundary Detection Challenge. The result is copied from the technical report and is reported on a subset of Kineics-GEBD validation set. As the solution incorporates optical flow features and model ensemble to boost performance, we do not directly compare our method with it.} 
\end{tablenotes}
\end{minipage}
\end{table*}
}

\newcommand{\TableSceneSOTA}{
\begin{table}[ht]
	\caption{Comparison with the scene-level state-of-the-art methods in terms of AP and $M_{iou}$ on MovieScenes and MovieNet.}
	\centering
     \setlength{\tabcolsep}{8pt}
    \renewcommand{\arraystretch}{1.1}
		{
		    \begin{threeparttable}
			\begin{tabular}{cccc}
				\toprule
				Method  & Modality          
				& AP ($\uparrow$) & $M_{iou}$ ($\uparrow$) \\
				\midrule
				GraphCut~\cite{DBLP:conf/cvpr/RasheedS03} & Visual
				& 14.1 & 29.7 \\
				SCSA~\cite{DBLP:journals/tmm/ChasanisLG09} & Visual
				& 14.7 & 30.5 \\
				DP~\cite{DBLP:conf/icmcs/HanW11}  & Visual
				& 15.5 & 32.0 \\
				Grouping~\cite{DBLP:conf/ism/RotmanPA16} & Visual
				& 17.6 & 33.1 \\
				StoryGraph~\cite{DBLP:conf/cvpr/TapaswiBS14} & Visual
				& 25.1 & 35.7 \\
				Siamese~\cite{DBLP:conf/mm/BaraldiGC15} & Visual
				& 28.1 & \text36.0 \\
				\multirow{2}{*}{LGSS~\cite{DBLP:conf/cvpr/RaoXXXHZL20}} & Visual Audio 
				& \multirow{2}{*}{47.1} & \multirow{2}{*}{48.8} \\
				 &  Actor Action & &  \\
                    \midrule
				 Temporal Perceiver  &  Visual
				& {\bf 51.9$\pm$2.4 } & {\bf 53.1$\pm$1.5} \\
			Temporal Perceiver\footnotemark[1] & Visual
				& \textbf{53.3} & \textbf{53.2} \\
				\bottomrule
			\end{tabular}
                \end{threeparttable}
        }
	\label{table:compare_SBD}
    \begin{tablenotes}
    \item \textit{\footnotemark[1] indicate results evaluated on MovieNet. } 
    \end{tablenotes}
\end{table}
}

\newcommand{\TableAblation}{
\begin{table*}

\caption{\textbf{Ablative Experiments} of Temporal Perceiver on different  choices of alignment implementations, coherence score implementations and input temporal length.}
\vspace{-1em}
\begin{center}
\setlength{\tabcolsep}{6pt}
 \renewcommand{\arraystretch}{1.1}
\begin{tabular}{cccccccccc}
\toprule
 &
   &
  \multicolumn{2}{c}{Alignment Implements} &
  \multicolumn{2}{c}{Coherence Implements} &
  \multicolumn{3}{c}{Input Temporal Length} &
   \\ \cmidrule(lr{0.5em}){3-4}\cmidrule(l{0.5em}r{0.5em}){5-6}\cmidrule(l{0.5em}r){7-9}
\multirow{-2}{*}{Benchmarks} &
  \multirow{-2}{*}{Metrics} &
  \multirow{-1}{*}{None} &
  \multirow{-1}{*}{Concatenation} &
  Gaussian at training &
  Likeliness at training &
  \multirow{-1}{*}{ 0.5N} &
  \multirow{-1}{*}{ 2N} &
  \multirow{-1}{*}{ 3N} &
  \multirow{-2}{*}{\begin{tabular}[c]{@{}c@{}}Temporal \\ Perceiver\end{tabular}} \\ \midrule
 &
  AP & 50.6& 49.6& 49.5& 50.5& 52.2&47.2&19.2&{\bf 53.3}
   \\
\multirow{-2}{*}{MovieNet} &
  $M_{IoU}$ &51.9&52.2&52.2 & 52.0& 52.0& 51.7& 38.4&{\bf 53.2}
   \\ \midrule
  &
  f1@0.05 & 73.7&73.5&74.7& 74.2& 69.4& 74.5& 33.8&{\bf 74.8}
  \\
\multirow{-2}{*}{Kinetics-GEBD} &
  avg-f1 & 83.3&82.2&84.6 & 82.8& 78.9&85.6 &49.7 & {\bf 86.0}
   \\ \bottomrule
\end{tabular}
\end{center}
\begin{tablenotes}
\item
\end{tablenotes}
\label{table:ablation}
\end{table*}
}

\newcommand{\TableTransformer}{
\begin{table*}[!t]
\caption{Quantitative Comparison with Transformer-based and Perceiver-based baselines.}
\setlength{\tabcolsep}{18pt}
\renewcommand{\arraystretch}{1.1}
\begin{threeparttable}
    \begin{tabular}{ccccccc}
            \toprule
              \multirow{2}{*}{Method} & {Feature } &{Alignment } &\multicolumn{2}{c}{MovieNet} & \multicolumn{2}{c}{Kinetics-GEBD} \\
              \cline{4-7}
              & Compression & Loss &AP&$M_{iou}$&f1@0.05&avg-f1 \\
              \midrule
              Sparse Transformer & $2\times$ down-sample & &46.2 &50.4 & 70.4 & 81.7\\
              Perceiver\footnotemark[2] & \cmark & &50.6 &51.9 & 73.7 & 83.3\\
              Temporal Perceiver &\cmark & \cmark & {\bf 53.3}& {\bf 53.2}&  {\bf 74.8} & {\bf 86.0}\\
              \bottomrule
    \end{tabular}
    \label{table:comparison_transformer}
    \begin{tablenotes}
    \item \footnotemark[2] We do not implement specific decoding queries or read-process sequence as in Perceiver and Perceiver IO in this variant.
    \end{tablenotes}
\end{threeparttable}
\end{table*}
}

\newcommand{\TableEfficiency}{
\begin{table}[ht]
	\caption{Efficiency Comparison with previous scene-level GBD method.}
        \vspace{-1em}
	\begin{center}
      \setlength{\tabcolsep}{17pt}
    \renewcommand{\arraystretch}{1.1}
	
			\begin{tabular}{ccc}
				\toprule
				Method  & ShotPS($\uparrow$)  & GFLOPs ($\downarrow$) \\
				\toprule
			    LGSS \cite{DBLP:conf/cvpr/RaoXXXHZL20} & 60.7  & 8141 \\
    			Vanilla Transformer & 371.1& 66 \\
			    Temporal Perceiver & \bf 377.9 & \bf 49 \\
				\bottomrule
			\end{tabular}
		
  \begin{tablenotes}
         \item \textit{The backbone network is excluded for all models in run-time.}
     \end{tablenotes}
	\end{center}
	\label{table:efficiency}
\end{table}
}

\begin{document}

\title{Temporal Perceiver: A General Architecture for Arbitrary Boundary Detection}

\author{Jing~Tan,
        Yuhong~Wang,
        Gangshan~Wu~\IEEEmembership{Member,~IEEE},
        Limin~Wang~\IEEEmembership{Member,~IEEE}
\IEEEcompsocitemizethanks{
\IEEEcompsocthanksitem J.Tan, Y.Wang, G.Wu and L.Wang are with the State Key Laboratory for Novel Software Technology, Nanjing University, Nanjing 210023, China.

(E-mail: \{jtan,yhwang24\}@smail.nju.edu.cn, \{gswu,lmwang\}@nju.edu.cn) }  

\thanks{Manuscript received April 19, 2005; revised August 26, 2015.}}

\markboth{Journal of \LaTeX\ Class Files,~Vol.~14, No.~8, August~2015}%
{Shell \MakeLowercase{\textit{et al.}}: Bare Advanced Demo of IEEEtran.cls for IEEE Computer Society Journals}

\IEEEtitleabstractindextext{
\begin{abstract}
    Generic Boundary Detection (GBD) aims at locating the general boundaries that divide videos into semantically coherent and taxonomy-free units, and could serve as an important pre-processing step for long-form video understanding. Previous works often separately handle these different types of generic boundaries with specific designs of deep networks from simple CNN to LSTM. Instead, in this paper, 
    we present {\em Temporal Perceiver}, a general architecture with Transformer, offering a unified solution to the detection of arbitrary generic boundaries, ranging from shot-level, event-level, to scene-level GBDs. The core design is to introduce a small set of latent feature queries as anchors to compress the redundant video input into a fixed dimension via cross-attention blocks. Thanks to this fixed number of latent units, it greatly reduces the quadratic complexity of attention operation to a linear form of input frames. Specifically, to explicitly leverage the temporal structure of videos, we construct two types of latent feature queries: boundary queries and context queries, which handle the semantic incoherence and coherence accordingly. Moreover, to guide the learning of latent feature queries, we propose an alignment loss on the cross-attention maps to explicitly encourage the boundary queries to attend on the top boundary candidates. Finally, we present a sparse detection head on the compressed representation, and directly output the final boundary detection results without any post-processing module. We test our Temporal Perceiver on a variety of GBD benchmarks. Our method obtains the state-of-the-art results on all benchmarks with RGB single-stream features: SoccerNet-v2~(81.9 percent average-mAP), Kinetics-GEBD~(86.0 percent average-f1), TAPOS~(73.2 percent average-f1), MovieScenes~(51.9 percent AP and 53.1 percent $M_{iou}$) and MovieNet~(53.3 percent AP and 53.2 percent $M_{iou}$), demonstrating the generalization ability of our Temporal Perceiver. 
\end{abstract}

}

\maketitle

\IEEEdisplaynontitleabstractindextext

\IEEEpeerreviewmaketitle

\ifCLASSOPTIONcompsoc
\IEEEraisesectionheading{\section{Introduction}\label{sec:introduction}}
\else
\section{Introduction}
\label{sec:introduction}
\fi

\IEEEPARstart{V}{ideo} content analysis~\cite{DBLP:conf/cvpr/CarreiraZ17,DBLP:conf/eccv/WangXW0LTG16,DBLP:journals/corr/abs-2203-12602,DBLP:conf/iccv/ZhiTWW21,DBLP:conf/eccv/LiW0W20,DBLP:conf/iccv/TengWLW21,DBLP:conf/iccv/JiDN21,DBLP:conf/cvpr/RaiCJDKI0N21,DBLP:conf/cvpr/JiK0N20} is a fundamental and important topic in computer vision due to the drastic growth of videos captured and shared online. Mainstream video understanding focuses on shot-form videos for action recognition~\cite{DBLP:conf/nips/SimonyanZ14,DBLP:conf/cvpr/WangL0G18,DBLP:conf/iccv/Feichtenhofer0M19,DBLP:conf/iccv/ZhiT0W21,DBLP:conf/cvpr/LiJSZKW20,DBLP:conf/cvpr/0002TJW21}, action detection~\cite{DBLP:conf/cvpr/ChaoVSRDS18,DBLP:conf/cvpr/Lin0LWTWLHF21,DBLP:conf/iccv/TanTWW21,Zhao_2022_CVPR,DBLP:conf/iccv/KalogeitonWFS17a,DBLP:conf/iccv/SinghSSTC17}, video retrieval~\cite{DBLP:conf/eccv/Gabeur0AS20,DBLP:conf/cvpr/LeiLZGBB021,Ge_2022_CVPR}, and video grounding~\cite{DBLP:conf/cvpr/ZhangDWWD19,DBLP:conf/aaai/ZhangPFL20,DBLP:conf/emnlp/XiaoCSZ021,DBLP:conf/aaai/00010WLW22,yang2022tubedetr}. Long-form videos, such as surveillance videos, movies, and recordings of sport events, contain hours of content with rich semantics. These long-form videos remain under-explored in previous literature due to the deficit in temporal segmentation of long videos to different levels. To facilitate long-form video understanding, we perform research to bridge the gap between long-form and short-form video understanding, by segmenting long videos into a series of shorter, meaningful pieces, as a basic pre-processing block. 

\begin{figure}[t]
  \centering
   \includegraphics[scale=0.4]{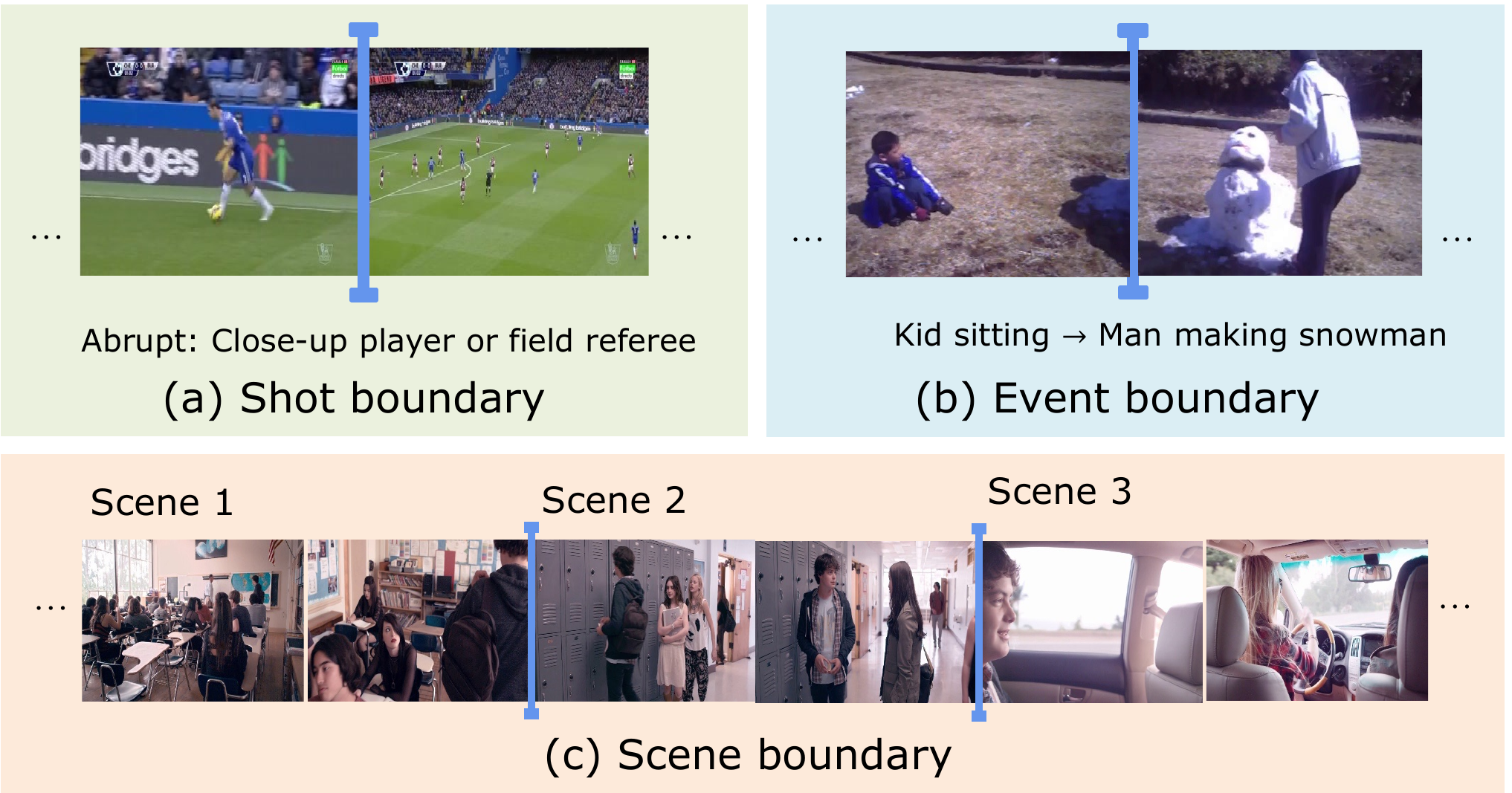}
   \caption{Examples of generic boundaries, from shot-level, event-level to scene-level boundary.}
   \label{fig:overview}
\end{figure}

We study the problem of general generic boundary detection, which aims to localize the temporal location of generic boundaries in long-form videos. 
The core concept of this task is {\bf generic boundary}~\cite{DBLP:conf/iccv/ShouLWGF21}, which is a specific kind of temporal boundary that {\em emerges only from semantic incoherence}. 
Different from the well-studied action instance boundaries of limited target classes~\cite{DBLP:conf/iccv/LinLLDW19,DBLP:conf/aaai/GaoSWLYGZ20,DBLP:conf/eccv/WangGWLW20,DBLP:conf/eccv/BaiWTYLL20,DBLP:conf/cvpr/XuZRTG20}, generic boundaries are {\em not specific to any pre-defined semantic category} and {\em can indicate the temporal structure of videos without any semantic bias}. The concept was explored in~\cite{DBLP:conf/iccv/ShouLWGF21} in regard to event-level generic boundary. We extend this study to arbitrary conditions including shot-level boundary in soccer matches~\cite{DBLP:conf/cvpr/DeliegeCGSDNGMD21} and scene-level boundary in movies~\cite{DBLP:conf/eccv/HuangXRWL20}.
\cref{fig:overview} provides examples of class-agnostic generic boundaries from shot-level, event-level to scene-level instances. To detect such diverse temporal boundaries, different levels of information are expected for capturing temporal structure and context at different scales.

Current research on generic boundary detection is separately studied in different tasks based on their boundary definition and granularity. For example, camera shot segmentation~\cite{DBLP:conf/cvpr/DeliegeCGSDNGMD21} targets at shot-level boundaries (\cref{fig:overview}a) that are rapid transitions between camera shots. Generic event boundary detection (GEBD)~\cite{DBLP:conf/iccv/ShouLWGF21} aims to locate event-level generic boundaries (\cref{fig:overview}b) which are moments where the action/subject/environment changes. Scene segmentation~\cite{DBLP:conf/eccv/HuangXRWL20} tries to detect scene-level generic boundaries (\cref{fig:overview}c), which are transits between movie scenes, indicating the high-level plot twists.
Previous methods~\cite{scikit,DBLP:conf/cvpr/RaoXXXHZL20,DBLP:conf/cvpr/ChenNFZBH21,DBLP:conf/iccv/ShouLWGF21} mainly focus on building informative feature representations carefully designed for the specific boundary type and cast the boundary detection problem as a dense prediction task. Shot boundary depends on low-level appearance and \cite{scikit} directly used local RGB difference to predict shot boundary. Event boundary is based on motion semantics and \cite{DBLP:conf/iccv/ShouLWGF21} used CNNs in local temporal window to address event boundary. Scene boundary is determined by holistic scene understanding and ~\cite{DBLP:conf/cvpr/RaoXXXHZL20} resorted to LSTM to capture long-range temporal context for scene boundary. Moreover, to produce the final boundary detection, these methods often resort to sophisticated post-processing techniques to remove duplicate false positives. The complicated design and post-processing is highly correlated with specific boundary type and prevent them from generalizing well to different types of generic boundary detection.

Instead, we argue that in spite of the difference in boundary granularity, these tasks share the similar semantic structure and pose similar requirement. Therefore, a natural question arises: {\em whether we can address these different kinds of generic boundary detection in a general and unified architecture?} To this end, in this paper, we unify the detection of different kinds of boundaries within a simple sparse detection framework. We introduce the {\em Temporal Perceiver}, a general and efficient framework designed to detect arbitrary kinds of boundaries using a single Transformer-based architecture. Attention modules are flexible architectural building blocks, but also require computational cost quadratically with the input dimension. Based on the key observation that there exists high redundancy among frames in each segment, we figure out to compress the long-form video before applying attention operations. Specifically, our core contribution is to introduce a small set of latent units as anchors to adaptively compress the long video sequence into a fixed dimension. These latent units form a cross-attention block to squeeze the original video input into a latent feature space with fixed dimension. This step reduces the computational cost of attention operations to be linear complexity of temporal duration. The compression rate varies according to different levels of redundancy in each benchmark. We show in experiments that the input features can be compressed into $\sim 50 \%$ for most benchmarks and the largest compression is up to $68\%$ for SoccerNet-v2. 

To improve the compression effectiveness of latent units, we propose two specific designs on their form and training strategy, respectively. Naturally, videos can be viewed a series of semantically coherent segments divided by generic boundaries. A good compression for GBD is expected to only skip these redundant frames inside the segments yet still retain the informative ones around the boundaries. First, to meet this requirement, our latent units is explicitly composed of boundary queries and context queries. The boundary queries aim to aggregate the important semantic incoherence regions across boundaries in a one-on-one manner, while the context queries are learned to cluster the coherence regions into a few semantic centers to suppress video redundancy. In practice, we compute an array of coherence scores on every temporal location to characterize the coherence structure. 
Second, to utilize this coherence score for effective compression, we propose a new alignment loss to guide the training on cross-attention maps and make this compression process more informative. This novel loss will endow our Temporal Perceiver with the awareness of video-specific temporal structure during latent unit learning, which turns out to be helpful for speeding up model convergence as well as boosting detection performance. 
Finally, we place a transformer decoder on top to perform sparse detection of generic boundaries with a set of learnable proposals. Due to the nature of sparse detection and strict matching criteria during training, our method is free of complicated post-processing and can be generalized to different types of generic boundary detection.

We perform experiments on five popular benchmarks across the shot-level, event-level and scene-level GBD tasks: SoccerNet-v2~\cite{DBLP:conf/cvpr/DeliegeCGSDNGMD21}, Kinetics-GEBD~\cite{DBLP:conf/iccv/ShouLWGF21}, TAPOS~\cite{DBLP:conf/cvpr/ShaoZDL20}, MovieScenes~\cite{DBLP:conf/cvpr/RaoXXXHZL20} and MovieNet~\cite{DBLP:conf/eccv/HuangXRWL20}, to verify the effectiveness of Temporal Perceiver for arbitrary generic boundary detection. In practice, Temporal Perceiver employs image encoder ResNet50~\cite{DBLP:conf/cvpr/HeZRS16} as backbone network by default for fair comparison with previous literature. Our method achieves the state-of-the-art performance on all benchmarks with remarkable improvements with RGB inputs only. 
Additionally, we test the performance of Temporal Perceiver with the IG-65M~\cite{DBLP:conf/cvpr/GhadiyaramTM19} pre-trained video encoder CSN~\cite{DBLP:conf/iccv/TranWFT19} as backbone on Kinetics-GEBD val set to demonstrate our performance upper-bound.
We also visualize qualitative results for better understanding of our models. We hope these quantitative results can provide more insightful analysis on our modeling details of our Temporal Perceiver.

In summary, our contributions are summarized as follows: 1) We present {\em Temporal Perceiver}, a general architecture to tackle generic boundary detection in long-form videos, providing a unified solution to the detection of the arbitrary boundary with Transformer. 2) To tackle the temporal redundancy of long videos and reduce the model complexity, we introduce a small set of latent queries for feature compression via cross-attention mechanism. 3) To improve the compression effectiveness of latent units, we devise customized the design on their form and training strategy, respectively.  4) Extensive experiments demonstrate that our method outperforms the existing state-of-the-art methods with RGB features only on shot-level, event-level, and scene-level generic boundary benchmarks, demonstrating its generalization ability on generic boundary detection.

\section{Related Work}
\label{sec:relatedwork}
In this section, we first discuss previous works in generic boundary detection, with a particular focus on shot-level, event-level and scene-level boundary detection. After that, we mention several related Visual Transformer literature that tackle the quadratic encoder complexity. 

\subsection{Generic Boundary Detection}
Generic boundary detection is termed for unified detection of arbitrary temporal generic boundaries. In this research, we particularly study GBD at shot-level (also termed as camera shot segmentation), event-level (GEBD), and scene-level (movie scene segmentation). 

Previous literature used specific hand-crafted techniques to solve different types of boundaries. Methods from Scikit~\cite{scikit} and PySceneDetect~\cite{PyScenedetect} libraries achieved superior performance in shot boundary detection based on frame variations in color or intensity. Previous GEBD methods~\cite{DBLP:conf/iccv/ShouLWGF21,DBLP:conf/cvpr/ShaoZDL20,DBLP:conf/eccv/LeaRVH16,DBLP:conf/eccv/LinZSWY18,DBLP:conf/iccv/LinLLDW19,DBLP:conf/cvpr/DingX18,DBLP:conf/eccv/HuangFN16} focused on building representations specifically designed for event-level boundaries and exploited dense prediction paradigm with post-processing. Temporal Convolution Network (TCN)~\cite{DBLP:conf/eccv/LeaRVH16,DBLP:conf/eccv/LinZSWY18} used a two-layer temporal convolution network to densely predict the confidence score for each frame.  PC~\cite{DBLP:conf/iccv/ShouLWGF21} explored the local temporal dependency by generating past and feature features for each temporal location for the final dense confidence estimation. These methods used CNN in a local temporal window to address event boundary and depended heavily on post-processing to remove duplicates. Scene segmentation attracts academic attention from both supervised~\cite{DBLP:conf/ism/RotmanPA16,DBLP:conf/mm/BaraldiGC15,DBLP:conf/cvpr/RaoXXXHZL20,DBLP:journals/tcsv/LiuKBP21} and unsupervised~\cite{DBLP:conf/icmcs/RuiHM98,DBLP:conf/cvpr/RasheedS03,DBLP:journals/tmm/ChasanisLG09,DBLP:conf/icmcs/HanW11,DBLP:conf/cvpr/TapaswiBS14,DBLP:conf/cvpr/ChenNFZBH21} research. The pioneering method LGSS~\cite{DBLP:conf/cvpr/RaoXXXHZL20} resorted to LSTM to capture long-range temporal context for scene boundary detection and manually designed a dynamic programming algorithm for post-processing. No previous work has addressed these boundaries in a single and unified framework. In contrast, our temporal perceiver presents a general solution with global view to handle these different types of boundaries without any specific design.

\subsection{Visual Transformers}
Inspired by the great advances of Transformer architecture~\cite{DBLP:conf/nips/VaswaniSPUJGKP17} in NLP field~\cite{DBLP:conf/naacl/DevlinCLT19}, researchers have explored the application of such architecture in visual tasks~\cite{DBLP:conf/eccv/CarionMSUKZ20,DBLP:conf/iclr/DosovitskiyB0WZ21,DBLP:conf/iccv/LiuLCHWZLG21,DBLP:journals/corr/abs-2106-13230,DBLP:conf/icml/TouvronCDMSJ21,DBLP:conf/iccv/ArnabDHSLS21,DBLP:conf/iccv/TanTWW21}. DETR~\cite{DBLP:conf/eccv/CarionMSUKZ20} was the first to successfully utilize a Transformer encoder-decoder structure in object detection task and achieves on par performance with its highly optimized CNN-based counterparts. Despite its success, DETR suffered from the quadratic memory complexity in encoder self-attention layers and therefore had limitations with large input sequences. Deformable DETR~\cite{DBLP:conf/iclr/ZhuSLLWD21} replaced the global and dense self-attention in encoder with a set of sparsely sampled reference points and deformable attention mechanism. PnP-DETR~\cite{DBLP:conf/iccv/WangYCFY21} performed differentiable sparse sampling on input sequences and reduced the input length to Transformer encoder. RTD~\cite{DBLP:conf/iccv/TanTWW21} extended the idea of DETR to the video domain for temporal action localization with several important improvements. In our Temporal Perceiver, we present a new architecture composed of two transformer decoders. Our first decoder is to re-purpose the Transformer decoder as an encoder for temporal feature compression in video domain. This decoder-as-encoder structure squeezes temporal redundant input into a small set of feature queries for subsequent processing. Our second decoder is similar to the work of DETR and RTD, but designed for a different task (i.e. generic boundary detection).

Recently, Jaegle \etal proposed the Perceiver ~\cite{DBLP:conf/icml/JaegleGBVZC21} and Perceiver IO~\cite{DBLP:journals/corr/abs-2107-14795} architecture, which formulates prediction tasks across modalities with a general ``read-process-write" perception paradigm. They read large inputs into small latent dimension through cross attention blocks, process the latent queries with stacked self-attention blocks, and outputs prediction results with cross attention blocks. 
In fact, we developed our method independently, but lately found our idea similar to the Perceiver architecture in compressing large input to small latent space. However, our Temporal Perceiver still has several important differences with Perceiver or Perceiver IO. First, our Temporal Perceiver deals with a temporal {\em detection} problem in videos using a sparse detection head, while Perceiver and Perceiver IO handles with {\em classification} and {\em dense prediction} problems, respectively. More importantly, we explicitly leverage the temporal structure for the design of latent units and their training strategy. Our explicit alignment loss will help to speed up the convergence of latent units and contribute to a better detection performance. Finally, in our Temporal Perceiver, the specific design of network architecture is different. The same number of layers of self-attention and cross-attention blocks are coupled and stacked alternatively to compress video features, while Perceiver and Perceiver IO use less cross-attention blocks and more self-attention blocks in separate reading and processing steps, respectively. We discuss the difference in details between our method and the Perceivers in \cref{sec:discussion}.

\begin{figure*}[ht]
\begin{center}
\includegraphics[scale=0.75]{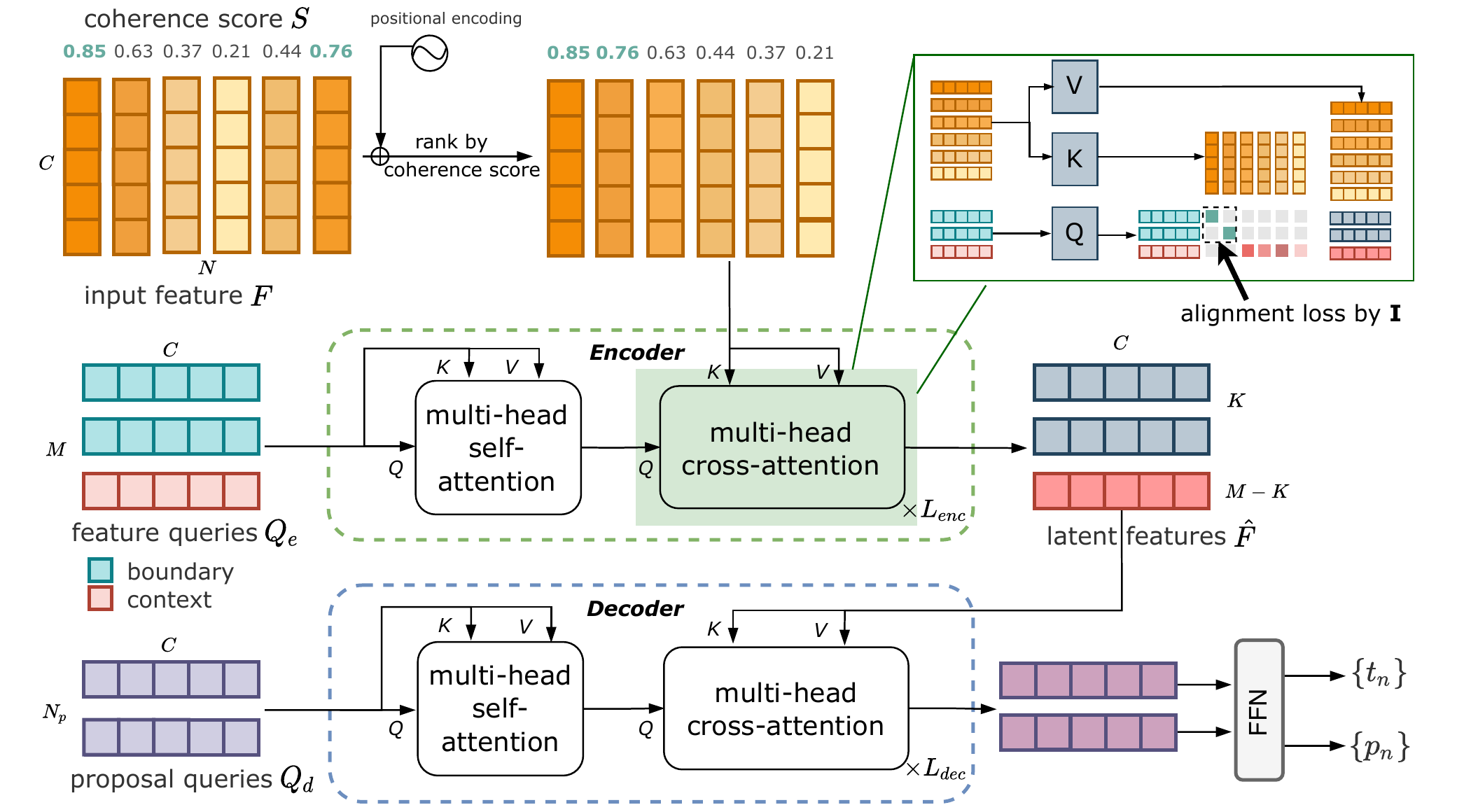}
\end{center}
   \caption{{The overall pipeline of Temporal Perceiver.} The input to our model is video features and an array of coherence scores, both of length $N$. Added with sine positional encoding, the input video features are shuffled and ranked in the descending order of coherence score. Feature queries compress the input feature into latent dimension $M$ in the encoder cross-attention. An alignment loss is enforced on the cross-attention map to make sure that $K$ boundary queries preserve top-$K$ boundary features and $M-K$ context queries cluster context features into $M-K$ semantic centers. After the feature compression, the decoder with FFNs (Feed-forward Networks) employs $N_p$ proposal queries to decode boundary location and confidence from the compressed latent sequence. }
\label{fig:arch}
\end{figure*}

\section{Method}
\label{sec:method}
\subsection{Overview}
We propose the Temporal Perceiver based on Transformers, to efficiently locate arbitrary generic boundaries in long-form videos. Given an untrimmed video $X$, Temporal Perceiver generates a set of generic boundary proposals $\Psi = \{t_n\}_{n=1}^{N_d}$ to locate the generic boundaries $\hat{\Psi} = \{\hat{t}_n\}_{n=1}^{N_g}$ in video $X$, with $N_d$ and $N_g$ denote the number of detected boundary and boundary groundtruths, respectively. 

The main structure of Temporal Perceiver is depicted in \cref{fig:arch}. We take the frame-level {\bf RGB features} $F \in \mathbb{R}^{N\times C}$ and {\bf coherence scores} $S \in \mathbb{R}^{N \times 1}$ from backbone network as input, where $N$ is the frame number and $C$ is feature dimension. A cross-attention based Transformer {\em encoder} $\mathbb{E}$ and a set of learnable {\bf feature queries} (latent units) $Q_e \in \mathbb{R}^{ M \times C}$, distill and transform the input sequence into a latent feature space of reduced dimension $\hat{F} \in \mathbb{R}^{M \times C}$, where $M$ is the number of latent units. The latent unit number is smaller than the original temporal dimension and each unit acts as an anchor to cluster the temporal information in a global view. These latent units could be supervised by an explicit alignment loss.
Based on this compressed representation, we use another Transformer {\em decoder} $\mathbb{D}$ and a set of learnable {\bf proposal queries} $Q_d \in \mathbb{R}^{N_p \times C} $ to directly  produce generic boundary representations. Finally, feed forward networks (FFNs) are used as localization head and scoring head to give the final predictions.
Our Temporal Perceiver yields a simple and end-to-end temporal detection framework without any post-processing technique, which makes no specific assumption about the boundary type and can be generalized to arbitrary generic boundary detection (GBD).

\begin{figure}[t]
  \centering
   \includegraphics[scale=0.6]{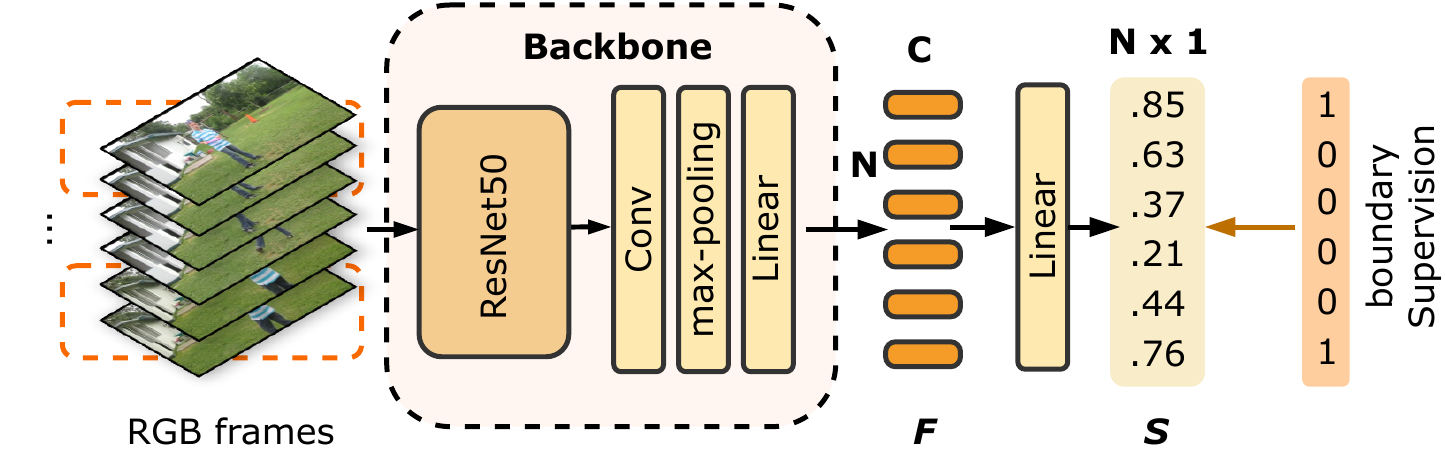}
   \caption{{An illustration of the backbone module.} The backbone are trained on benchmarks with dense boundary supervision. Video features are extracted after the max pooling on each snippet. The coherence scores are predicted from the final linear layer.}
   \label{fig:backbone}
\end{figure}

\subsection{Feature Encoding and Coherence Score}
As illustrated in \cref{fig:backbone}, we adopt ResNet50~\cite{DBLP:conf/cvpr/HeZRS16} pre-trained on ImageNet~\cite{DBLP:conf/cvpr/DengDSLL009} as backbone to extract appearance features $F = \{f_i\}_{i=1}^N $ and coherence scores $S = \{s_i\}_{i=1}^N$ from RGB frames. Specifically, we first densely sample snippets from videos. Each snippet contains $k$ frames and is sampled at stride $\tau$. Given a snippet $i$, we concatenate its $k$ feature vectors encoded by backbone, and then perform temporal convolution, max pooling and linear layer upon the feature vectors. The output feature $f_i$ is extracted after max-pooling layer. The final linear layer outputs a classification score $s_i$ to represent the probability that its center frame $i$ is boundary. This classification score is the coherence score used in Temporal Perceiver. The backbone network is trained with the dense boundary groundtruth on the benchmarks in advance.

\subsection{Temporal Perceiver}
The Temporal Perceiver architecture is formulated as an encoder-decoder framework and built upon the Transformer. The encoder, which is a re-purposed Transformer decoder of $L_{enc}$ layers, takes in a set of feature queries and compresses the video feature into a reduced sequence of latent representations. The decoder, which is a vanilla Transformer decoder of $L_{dec}$ layers, directly output final predictions with a set of proposal queries from the compressed feature. The input features of length $N$ are down-scaled into fixed length $M$ via the cross-attention operations in encoder $(M < N)$. 
Compared with the original Transformer encoder of fully self-attention, our Temporal Perceiver complexity is reduced from $O(N^2)$ to $O(NM+M^2)$, resulting in linear complexity of $N$ with a relatively small $M$.

\subsubsection{Encoder: Temporal Compression via Latent Units}
Videos can be divided into two kinds of regions in time, namely boundary regions and context regions (i.e., within each segment). Boundary region refers to the temporal interval around the generic boundary with gradual temporal transition, while context region refers to more central regions within coherent segments.
For better localization performance, our main focus is on boundary regions. Context regions are highly redundant and could be compressed into shorter sequences for efficiency. 
To this end, we build two types of feature queries (boundary queries and context queries) to attend on semantic incoherent (boundary) and coherent (context) regions accordingly. 

Specifically, we distinguish the boundary region and the context region based on dense coherence scores $S$ obtained from backbone. These coherence scores are estimates of the likelihood that the corresponding location is a generic boundary. We sort the input features according to the descending order of coherence scores and obtain the sorted permutation $\pi$.
The top $K$ locations with highest coherence scores are taken as boundary regions
\begin{equation}
    F_b = \{ f_{\pi_n} | n = 1,...,K \}.
    \label{equ:ranking}
\end{equation}
The rest as context regions $F_c = F~\symbol{92}~F_b$.
We allocate $M$ latent feature queries to $K$ boundary queries $Q_b$ and $M-K$ context queries $Q_c$. These latent queries $Q_e = [Q_b, Q_c]$  are learnable embeddings shared among all videos in the dataset. They are randomly initialized and trained jointly with network weights. 
Self-attention in encoder models the global dependency for feature queries, adding pair-wise feature interactions throughout the compression. 
Through cross-attention layers, $K$ boundary queries $Q_b$ handle the features of boundary regions in a one-on-one manner, and $M-K$ context queries $Q_c$ flexibly cluster the vast context regions into a few contextual centers. The overall encoder of Temporal Perceiver is as follows:
\begin{multline}
    \mathrm{Encoder} \ \ \mathbb{E}: \ \ \hat{F} = \\ \mathrm{Cross\_attention}([F_b, F_c],~\mathrm{Self\_attention}(Q_e)),
\end{multline}
where $\hat{F} \in \mathbb{R}^{M \times C}$ is the compressed video representation for subsequent sparse detection in decoder. 

\textit{Alignment Constraint on Cross-attention.}
To guide the boundary feature queries to attend the actual video boundaries, we incorporate alignment constraint on its cross-attention maps. As we have re-organized the feature sequence according to its coherence score in the ranking step (in \cref{equ:ranking}), we could simply introduce an identity matrix based alignment loss on the last cross-attention map of boundary queries. In this sense, each boundary query is enforced to only attend one of top $K$ locations with highest coherence score. For context queries, we allow them to flexibly attend context information without any constraints.
With this new training strategy, we find our feature queries can speed up the model convergence and generate more stable compressed feature representations corresponding to both boundaries and context.

\textit{Coherence Score Formulation.}
\label{sec:coherence}
The coherence structure of videos serves as the important prior for the novel improvements discussed above over feature compression. In practice, we devise a coherence score at each temporal location as a metric to tell whether this location is in the boundary or context regions. We study different formulations of coherence scores, described as follows:
\begin{itemize}
    \item \textbf{Summed Gaussian kernels} centered at each groundtruth generic boundary location. Summed gaussian kernels describe the precise distribution of boundary and context regions for training samples. However, it also creates gap for training and inference as inference samples do not have access to the groundtruth based gaussian kernels.  
    \item {\bf Learned boundary likeliness} are predictions of the boundary likeliness at each temporal location. 
    These are softer and more consistent boundary estimations that are available at both training and inference. 
\end{itemize}
We compare the performance of summed Gaussian kernels at training, learned boundary likeliness at training and learned boundary likeliness at both training and inference in ablation studies. According to empirical results, we set learned likeliness at both training and inference as the default coherence scores.

\subsubsection{Decoder: Sparse Detection via Proposal Queries}
We follow the direct proposal generation pipeline in~\cite{DBLP:conf/eccv/CarionMSUKZ20,DBLP:conf/iccv/TanTWW21} to generate boundary prediction with Transformer decoders and a set of learnable proposal queries. Throughout the stacked self-attention layers and cross-attention layers, the pair-wise dependencies are modeled for proposals via self-attention to avoid duplicates, and the full sequence of reduced features are attended to via cross-attention to generate proposal embedding. Specifically, the sparse detection is defined as follows:
\begin{multline}
    \mathrm{Decoder} \ \ \mathbb{D}: \ \ \Phi = \\ \mathrm{Cross\_attention}(\hat{F},~\mathrm{Self\_attention}(Q_d)),
\end{multline}
where $\Phi \in \mathbb{R}^{N_p \times C}$ is the decoded representations for all proposal queires.

The final predictions are decoded from proposal query embedding via a two-branch head architecture. The localization head uses a three-layer MLP to predict the boundary location, and the classification head uses a fully-connected layer followed by sigmoid function to get confidence scores that effectively distinguish boundaries. 

\subsubsection{Discussion with Perceiver and Perceiver IO}
\label{sec:discussion}
In spirit, our work is similar to the recent Perceiver~\cite{DBLP:conf/icml/JaegleGBVZC21} and Perceiver IO~\cite{DBLP:journals/corr/abs-2107-14795} architectures. Although we both utilize latent queries for input compression, our method is different from the Perceivers in several important aspects. 

First, the basic processing scheme and pipeline is different. Both Perceiver and Perceiver IO follow the general ``read-process-write'' scheme by designing separate compression and processing modules. In this decoupled architecture, it uses a asymmetric architecture where a cross attention block is equipped with multiple self-attention block. Instead, our Temporal Perceiver employs a coupled and progressive ``compression-process'' scheme by stacking multiple layers of encoders, each of which is compose of a cross-attention and self-attention block. Our Temporal Perceiver aims to gradually compress and transform the original video feature sequence in a joint manner. Therefore, the number of cross-attention and self-attention blocks in our Temporal Perceiver is the same. We analyze that this coupled and progressive compression and transformation process can enhance its effectiveness and also allows for deep supervision to guide the this process as discussed next.

Second, the training paradigm and loss is different. Both Perceiver and Perceiver IO fail to utilize the explicit loss to guide the training of latent units. They solely reply on the final target supervision such as classification and prediction loss. Yet, this blind learning process might lead to insufficient latent unit usage and inferior performance. Instead, we carefully consider the specific property of video data and the GBD task, and propose customized latent unit form and training strategy.  In particular, we explicitly divide the latent queries into two kinds (boundary and context) to handle the temporal redundancy of input video data. Meanwhile, we introduce a novel alignment loss on cross attention maps to guide the training of boundary queries, while Perceiver and Perceiver IO do not employ any auxiliary supervision. Ablative experiments show that the alignment loss helps to speed up the model convergence as well as improve model performance. 

Finally, the target problem and domain is different. Both Perceiver and Perceiver IO deal with the general classification and dense prediction in space. These tasks often require a classification head to generate a label or a dense prediction head to generate pixel-wise labels. Instead, our Temporal Perceiver handle the temporal detection problem in videos. We need to integrate our Temporal Perceiver with a sparse detection head to directly regress the location of generic boundary. In general, the detection task is more challenging than the classification and dense prediction task due to the large variance in its detected targets. Overall, direct application of Perceiver architecture is not sufficient for good boundary detection and our customized designs are effective with non-trivial performance gains in general generic boundary detection.
 
\subsection{Training}
 First, we divide videos features into fixed-length sequences via a sliding-window manner: short videos are zero-padded and long videos are divided into non-overlapping segments. Then, we filter out those windows with no groundtruth for training.
 
\subsubsection{Label Assignment}
Following the practice of \cite{DBLP:conf/eccv/CarionMSUKZ20}, We assign positive labels to predictions with groundtruths in a strict matching scheme. For video input $X$, the groundtruth set is denoted $\hat{\Psi} = \{\hat{t}_n\}_{n=1}^{N_g}$ and the prediction set is denoted $\Psi = \{t_n\}_{n=1}^{N_p}$. $N_p$ is assumed to be larger than $N_g$. The bipartite matcher searches for an optimal one-on-one matching between the two sets. The matcher minimizes the cost function via hungarian algorithm to seek the optimal matching $\sigma_*(\cdot)$, then assign positive labels to those predictions that are matched with groundtruths. The cost function is defined as:
\begin{equation}
    \mathcal{C} = \sum_{n:\sigma(n) \neq \emptyset} \alpha_{loc} \cdot \vert t_n -\hat{t}_{\sigma(n)}\vert - \alpha_{cls} \cdot {p_n},
\end{equation}
where $p_n$ denotes the boundary confidence score of proposal query $n$, $\sigma(\cdot)$ denotes a permutation for predictions to match with the groundtruths. $\alpha_{cls}$ and $\alpha_{loc}$ denote coefficients for classification error and localization error. 

\subsubsection{Loss Functions}
Conventionally, we define a localization loss and a classification loss to supervise the final predictions. The localization loss is formulated with $L_1$ loss on predicted boundary point $t_n$ and its matched groundtruth $\hat{t}_{\sigma(n)}$:
\begin{equation}
    L_{loc} = \frac{1}{N_{p,pos}}\sum_{n:\sigma_*(n)\neq \emptyset} L_1(t_n ,\hat{t}_{\sigma_*(n)}).
\label{loss:loc}
\end{equation}
The classification loss is defined with cross-entropy loss:
\begin{equation}
    L_{cls} = - \frac{1}{N_p}\sum_{n:\sigma_*(n)}
    (\hat{l}_{\sigma_*(n)}\log (p_{n}) + (1-\hat{l}_{\sigma_*(n)})\log(1-p_{n})),
\label{loss:cls}
\end{equation}
$\hat{l}_{\sigma_*(n)}$ denotes the matched binary groundtruth label for proposal query $n$.

To enforce additional constraint on feature compression in order to preserve the important boundary information, we add a diagonal alignment loss $L_{align}$ on cross-attention maps:
\begin{equation}
    L_{align} = - \log \frac{1}{K}\sum_{m=1}^{K}
    \mathbf{A}_{m,m},
\label{loss:diag}
\end{equation}
where $\mathbf{A}$ denotes the last-layer cross-attention map. It is noted that we use auxiliary loss with classification loss and localization loss, but not with the alignment loss.

The overall loss function $\mathcal{L}$ writes as follows:
\begin{equation}
   \mathcal{L} = \alpha_{loc} \cdot L_{loc} + \alpha_{cls} \cdot L_{cls} + \alpha_{align} \cdot L_{align},
\label{loss:overall}
\end{equation}
 where $\alpha_{align}$ is the alignment loss coefficient. We use the same set of classification and localization coefficients for the matching cost function and the loss function.
\subsection{Inference}
During inference, we directly and sparsely predict boundary locations as well as their confidence scores without the help of post-processing algorithms, such as watershed, dynamic programming or non-maximum suppression.

We use fully-connected layers as localization and classification heads for final predictions. These heads take in the proposal embedding and output boundary location $t_n$ and its corresponding confidence $p_n$ respectively. The generic boundary detection metrics require results in a hard-submission manner, without directly leveraging the confidence estimation performance. As a result, we threshold our predictions based on confidence threshold $\gamma$ to filter the most confident predictions for submissions.

\section{Experiments}
\label{sec:experiment}

In this section, we first introduce the datasets and implementation details of Temporal Perceiver for different levels of GBD. Then, we provide quantitative comparisons with the previous state-of-the-art models on all benchmarks. Additionally, we compare the model efficiency with dense transformer and previous works to show the benefits from feature compression. Finally, we provide detailed ablative analysis of all components in our model to proving the effectiveness of each component. We also compare our method with re-implemented transformer variants to demonstrate our improvement over transformer-based baselines.

\subsection{Datasets}
\textit{Shot-level Boundary Dataset.}
{SoccerNet-v2}~\cite{DBLP:conf/cvpr/DeliegeCGSDNGMD21} is a large-scale soccer video dataset, containing 500 untrimmed soccer game recordings with an average duration of 1.53 hour per game. Among them, 200 games are annotated with 158,493 shot boundaries for the task of camera shot boundary detection. Three types of shot transitions are annotated, including abrupt, fading and logo transition. The annotated games are divided into training, validation and testing set of 120, 40 and 40 recordings respectively.

\textit{Event-level Boundary Dataset.}
We evaluate the detection performance of event-level boundaries on two popular generic event boundary benchmarks: {Kinetics-GEBD} and {TAPOS}. 
Kinetics-GEBD\cite{DBLP:conf/iccv/ShouLWGF21} is a recently proposed benchmark for GEBD task. The dataset is derived from Kinetics-400~\cite{DBLP:journals/corr/KayCSZHVVGBNSZ17} dataset. It contains 60K YouTube videos that depicts various human actions and is divided into Train, Val and Test set. The Train and Test set each contains 20K videos from Kinetics-400 Train Set, and the Val set contains 20K videos from Kinetics-400 Val Set. 
TAPOS~\cite{DBLP:conf/cvpr/ShaoZDL20} contains 16,294 valid instances from Olympics sport videos. All instances are split into train, validation and test sets, of sizes 13094, 1790, 1763, respectively. We trim the action instances as input video and detect the sub-action boundaries within the action. The benchmark is re-purposed for generic event boundary detection by Shou et al. \cite{DBLP:conf/iccv/ShouLWGF21}. Our model is evaluated on the validation set.

\TableShotSOTA

\textit{Scene-level Boundary Dataset.}
{MovieScenes}~\cite{DBLP:conf/cvpr/RaoXXXHZL20} is scene segmentation dataset containing 297 hours of 150 movies. Previous works divide the dataset into a 100-video training set, a 20-video validation set and a 30-video test set. { MovieNet}~\cite{DBLP:conf/eccv/HuangXRWL20} dataset expands upon MovieScenes and becomes another popular benchmark for scene segmentation. It contains 42K annotated scene segments from 318 movies, and it is divided into training, validation and testing sets with 190, 64 and 64 movies respectively.  We compare with previous works on MovieScenes and report ablation results on MovieNet validation set.

\subsection{Implementation details.}
We adopt ResNet50~\cite{DBLP:conf/cvpr/HeZRS16} for feature extraction, the sampling stride $\tau = 1$ for SoccerNet-v2, $\tau = 3$ for Kinetics-GEBD and TAPOS. Due to copyright, MovieNet only provides three frames from each shot, so the actual temporal strides between frames can be huge. We strategically take all frames into account and set the sampling stride $\tau$ to $1$, to mitigate large regression errors caused by large input stride. The local window size $k$ for frame-level feature extraction is 9 for SoccerNet-v2, 10 for Kinetics-GEBD and TAPOS, 12 for MovieNet.

We process input videos via a sliding-window mechanism for all datasets, the window size $N$ for SoccerNet-v2, Kinetics-GEBD and TAPOS is set to 100 and the window size for MovieNet is set to 30. Short videos are zero-padded to window size. We don't overlap windows, so the overlap ratio is 1 for both training and inference. The refined feature length $M$ is set to 32, 60, 60, 15 for SoccerNet-v2, Kinetics-GEBD, TAPOS and MovieNet; the number of boundary queries and context queries is 20 and 12 for SoccerNet-v2, 48 and 12 for Kinetics-GEBD and TAPOS, 10 and 5 for MovieNet. $L_{enc} = 6$ for Kinetics-GEBD and TAPOS, $3$ for MovieNet and SoccerNet. $L_{dec} = 6$ for all benchmarks.

The loss parameters $\alpha_{cls}$, $\alpha_{loc}$ and $\alpha_{align}$ are set to $2, 1, 1$ respectively. We use AdamW as optimizer. The learning rate is set to 2e-4 and the batch size is 64. The confidence threshold $\gamma$ for final submissions is set to $0.9$ for SoccerNet-v2, Kinetics-GEBD and TAPOS, $0.7$ for MovieNet.

\TableEventSOTA

\subsection{Comparison with the State of the Art}
\subsubsection{Shot Boundary Detection}
{\em Evaluation Metrics.}
We follow \cite{DBLP:conf/cvpr/DeliegeCGSDNGMD21} to use mAP metric with tolerance $\delta$ of 1 second for evaluation. A predicted boundary is positive if it falls within the given tolerance $\delta$ of a ground-truth timestamp.

{\em Comparisons with the state of the art.}
We train Temporal Perceiver on the training set, validate on the validation set to select the best model, then report the results on the testing set. 
As \cref{tab:camera-shots-results} illustrates, Temporal Perceiver outperforms the previous state-of-the-art method by $3.4\%$, generalizing well to shot-level generic boundary detection. 
Our model outperforms all previous methods on fading transitions with a large margin of $25\%$ and achieves competitive results on abrupt and logo transitions, thanks to the adaptive receptive field of attention mechanism.

\subsubsection{Event Boundary Detection}
{\em Evaluation metrics.}
We use f1 score under different Relative Distance thresholds for quality measurement. Relative Distance is the relative distance between the predicted and groundtruth boundary timestamps, divided by the duration of the corresponding video. Given a fixed Relative Distance, we use it as threshold to determine whether a boundary prediction is correct, then the f1 metrics can be computed. We mainly compare f1 score with 0.05 relative threshold and average f1 score. 

{\em Comparisons with the state of the art.}
In \cref{table:compare_GEBD}, we compare the results of ResNet50-based Temporal Perceiver with the state-of-the-art methods on Kinetics-GEBD and TAPOS for event-level boundary detection. The results show that our method comfortably outperforms the state-of-the-art method especially with smaller Rel.Dis. threshold (less error tolerance). 
We also provide results of Temporal Perceiver with CSN backbone on Kinetics-GEBD to show the performance upper-bound of our model. This also demonstrates that our model produces more precise and accurate boundary predictions than previous methods both with image encoder and video encoder backbone. 

\subsubsection{Scene Boundary Detection}

{\em Evaluation metrics.}
Following \cite{DBLP:conf/cvpr/RaoXXXHZL20}, we use Average Precision (AP) and $M_{iou}$ to evaluate the quality of detected scene boundaries, where $M_{iou}$ calculates the weighted sum of intersection-over-union of a detected scene with respect to its distance to the closest ground-truth scene.

{\em Comparisons with the state of the art.}
In \cref{table:compare_SBD}, we compare the results of Temporal Perceiver with the state-of-the-art methods on MovieScenes. The results of previous literature is quoted from \cite{DBLP:conf/cvpr/RaoXXXHZL20}. However, the exact data-split scheme in \cite{DBLP:conf/cvpr/RaoXXXHZL20} is not provided. Thus, we apply a 10-fold cross-validation on the dataset and report the mean and standard deviation of AP and $M_{iou}$. 

To compute Average Precision, we construct Gaussian distributions centered at our sparsely-predicted boundary locations to approximate dense score sequences for each video. AP is reported based on the approximated dense scores. Temporal Perceiver improves the AP by $4.8\%$ and $M_{iou}$ by $4.3\%$ over previous SOTA, using visual modality only. The remarkable improvement demonstrates the generalization ability and effectiveness of our model on generic boundary detection under various granularity.

\TableSceneSOTA

\TableEfficiency

\TableTransformer

\subsection{Efficiency Analysis}
We analyze Temporal Perceiver's advantage in efficiency in terms of Shot Per Second (SPS) in inference and FLOPs of feature encoders on MovieNet. We compare our model with the previous state-of-the-art method LGSS and a Transformer baseline method. The results are reported on one RTX 2080-Ti GPU. We take three random movies selected from MovieNet validation set (2275 shots in total) as input. \cref{table:efficiency} shows that our model is more efficient than LGSS and the vanilla Transformer baseline due to effective feature compression. Free of post-processing modules, Temporal Perceiver infers $6$ times faster with much less FLOPs than LGSS.

\begin{figure*}[!t]
\centering
\subfloat[The model size grows with the increase of latent dimension $M$, while the detection performance achieves the sweet spot at 50\% compression for scene-level boundaries. ]{\includegraphics[scale=0.4]{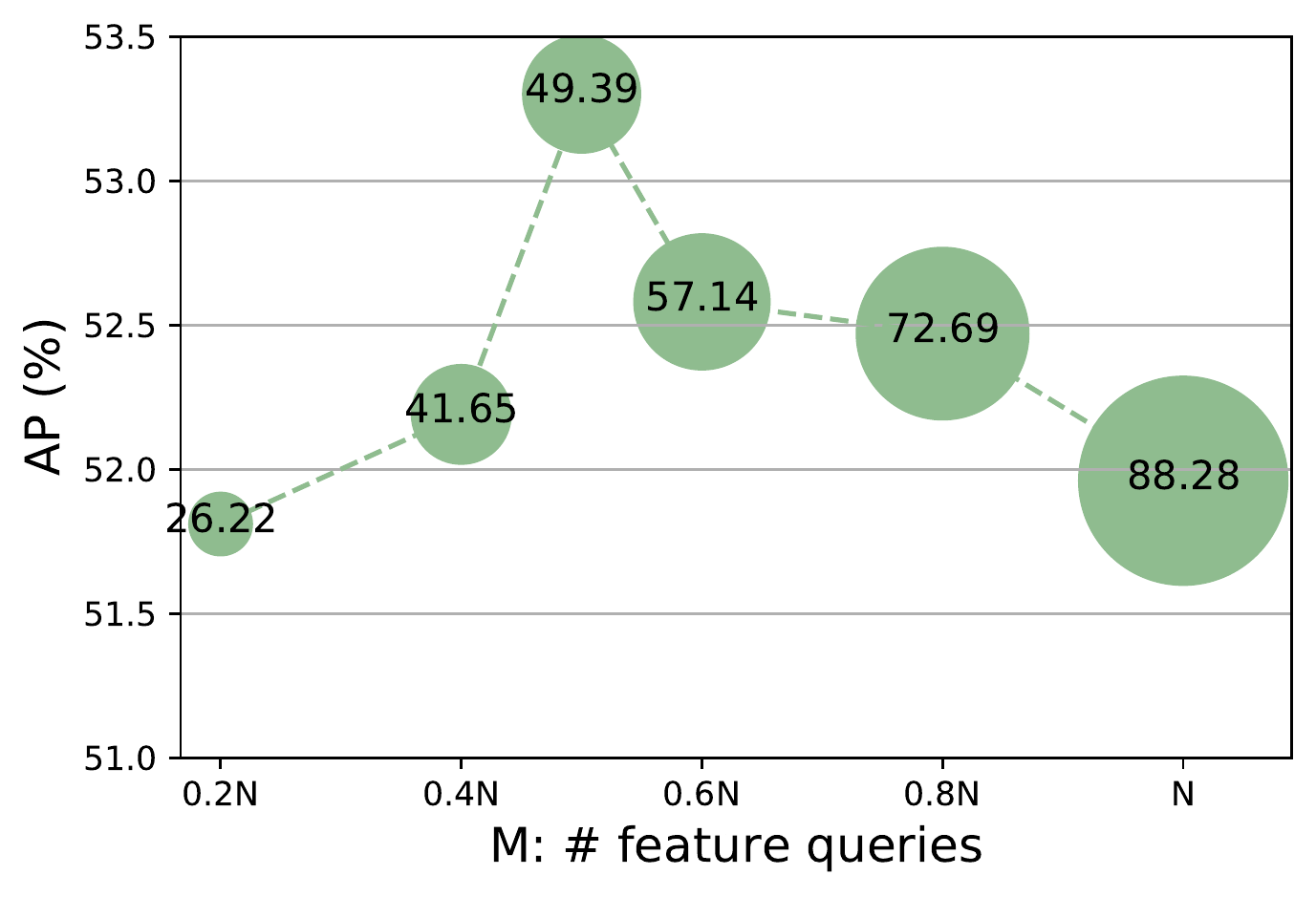}
\label{fig:abla_compression}}
\hspace{0.5em}
\subfloat[The performance is lower when there's only one type of queries and increases as the number of boundary and context queries gets balanced. However, it is better to have more boundary queries than context queries.]{\includegraphics[scale=0.4]{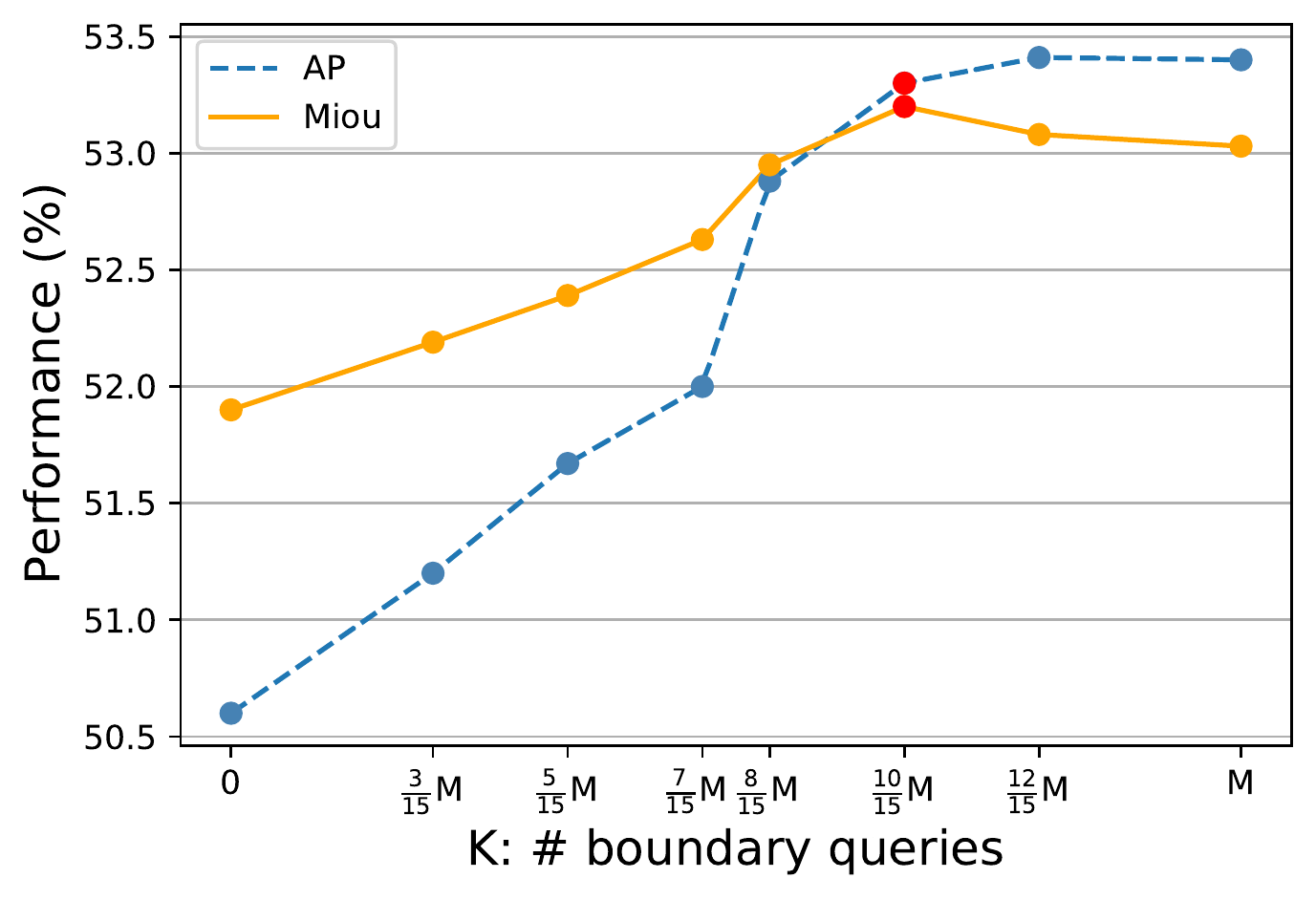}
\label{fig:balance}}
\hspace{0.5em}
\subfloat[Alignment loss helps the model to converge faster and also improve the performance by 1.1\% under f1@0.05 and 2.7\% under avg-f1.]{\includegraphics[scale=0.41]{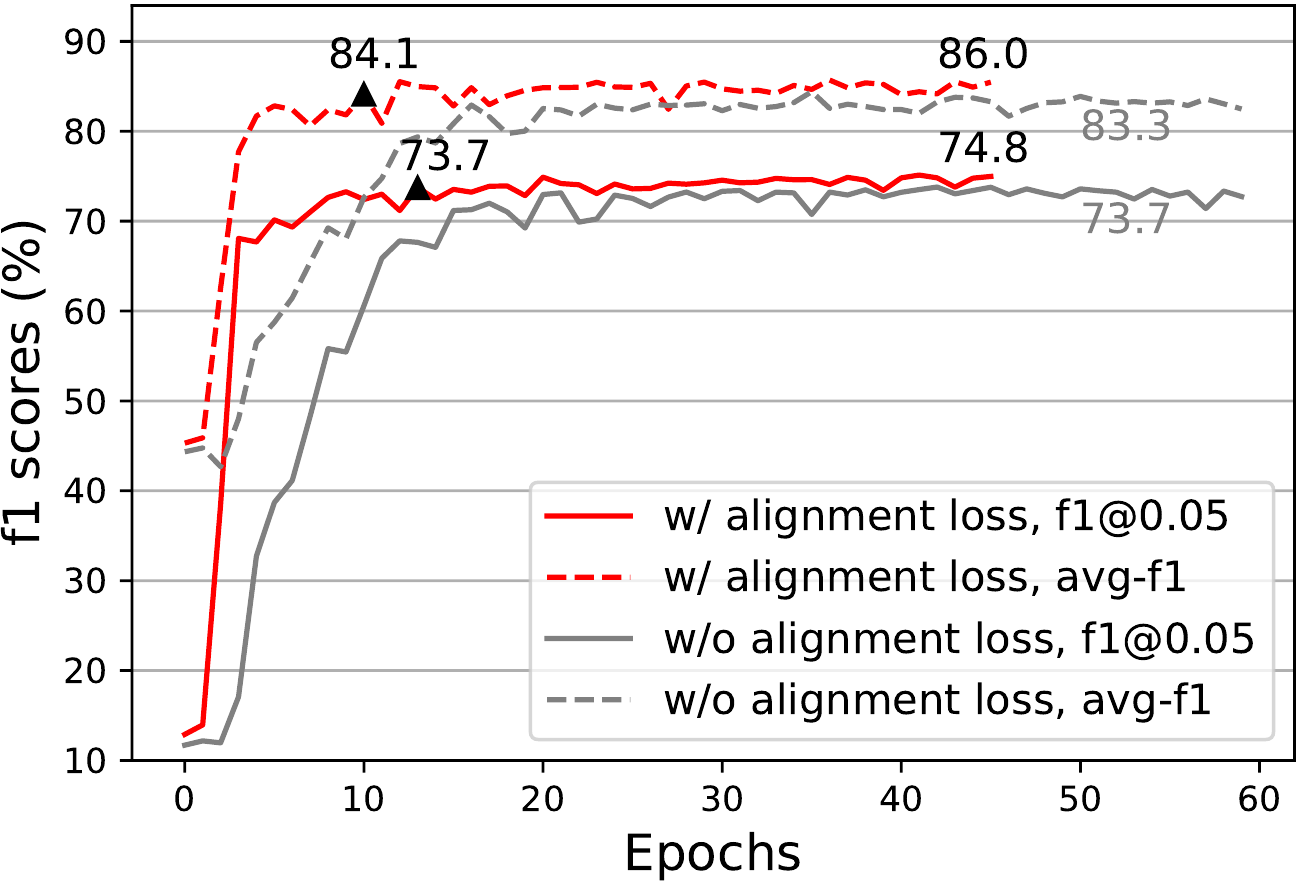}
\label{fig:compare_alignment}}
\caption{Qualitative ablation studies on (a) feature compression rate, (b) latent query construction and (c) model convergence with alignment loss.}
\label{fig_sim}
\end{figure*}

\TableAblation

\subsection{Ablation Study}
\textit{Comparison with Transformer-based baseline.}
 To achieve fair comparison and demonstrate the effectiveness of our model over Transformer architecture, we implement two baseline model: 1) sparse Transformer baseline with $2\times$ down-sample to compress feature before encoder processing; 2) Perceiver baseline consists of feature compression without coherence assumption and a Transformer decoder for boundary prediction. In \cref{table:comparison_transformer}, Sparse Transformer is clearly a strong baseline on both MovieNet and Kinetics-GEBD. Nevertheless, Perceiver baseline easily surpass Sparse Transformer baseline, verifying the effectiveness of learned compression over naive down-sample. Temporal Perceiver surpass both Sparse Transformer and Perceiver baselines with non-trivial gains on each metric, demonstrating the effectiveness of coherence structure modeling with specific query construction and the alignment loss.
 
\textit{Study on feature compression.}
 With latent queries and cross-attention mechanism, we reduce the temporal redundancy and model complexity via feature compression. \cref{fig:abla_compression} shows the AP and encoder GFLOPs on MovieNet under different settings of $M$. The point size and the captions represent encoder GFLOPs. The AP performance reaches the peak at $M=0.5N$, indicating the sweet spot for compression without severe information loss. It is interesting that lower compression also performs worse. We argue that while high compression could miss important content, at low compression the video redundancy would distort the quality of compressed features, making $M=0.5N$ a perfect solution for balance and efficiency. Compared with the zero-compression ($M=N$) setting, our design improves the performance by 1.34\% in AP and reduces the FLOPs by half.

\textit{Study on query construction.}
With specific latent query construction, we exploit the semantic structure of long-form videos and compress features into stable latent queries. The ratio  of boundary and context queries in latent query construction needs consideration. We test different number of boundary queries $K$ under a fixed $M = 0.5N$. \cref{fig:balance} shows that the detection performance is sub-optimal at both ends of the curve for $M_{iou}$. The performance drops drastically at extremely small $K$. We argue that when there are few boundary queries, the compressed features lack discriminative boundary information, explaining the weak performance. The performance is also weak at $K = M$ without context queries, demonstrating the complementary contributions of context information. Generally, it is observed that boundary information benefits the detection performance more than context information. We report with the setting $K=\frac{2}{3}M$ for the best performance based on empirical results.

\textit{Study on alignment loss.}
The alignment loss is proposed to guarantee the respective compression scheme for boundary and context queries. \cref{table:ablation} reports the ablation results on the alignment loss. With the additional alignment supervision, the f1@0.05 and avg-f1 performance increases by 1.1\% and 2.7\% on Kinetics-GEBD benchmark, and the $M_{iou}$ and AP performance increases by 1.3\% and 2.7\% on MovieNet. In addition, we plot the convergence curve on Kinetics-GEBD in \cref{fig:compare_alignment}, the results show that the model converges faster with the alignment loss and improves f1 score. 

In order to build stable boundary-context representations, the boundary queries attend to boundary features in a one-on-one manner via alignment supervision on the last layer of encoder cross-attention. For the design simplicity, one may ask: {\em why not directly use boundary features to replace learned boundary queries?} In \cref{table:ablation}, We compare the results of the alignment loss guided model and another variant model that discards the boundary queries and directly concatenates the boundary features with learned context centers. On both benchmarks, the performance of the direct concatenation variant is weaker than the alignment-guided model. We contend that boundary queries aggregate crucial context into feature embedding via layers of cross-attention (before the last layer) and self-attention.

\textit{Study on coherence scores.}
As discussed in \cref{sec:coherence}, we study two types of coherence scores implementation, namely summed gaussian kernels and learned boundary likeliness. In practice, we experiment with three different implementations in \cref{table:ablation}: 1) summed gaussian kernels used only at training; 2) learned boundary likeliness used only at training; 3) learned boundary likeliness used at both training and inference (also the default setting). Summed gaussian kernels has advantage over learned boundary likeliness at training due to the precise estimation of boundary/context regions. However, compared with gaussian kernels only at training, learned likeliness at both training and inference achieves better performance by reducing the learning difficulty for feature compression .

\begin{figure*}[!t]
\centering
\subfloat[Qualitative results of scene-level boundary detection on MovieNet. We select a movie clip to demonstrate the detection results. Compared to the previous supervised SOTA LGSS, our method achieves more accurate predictions with single-stream RGB input.]{\includegraphics[scale=0.6]{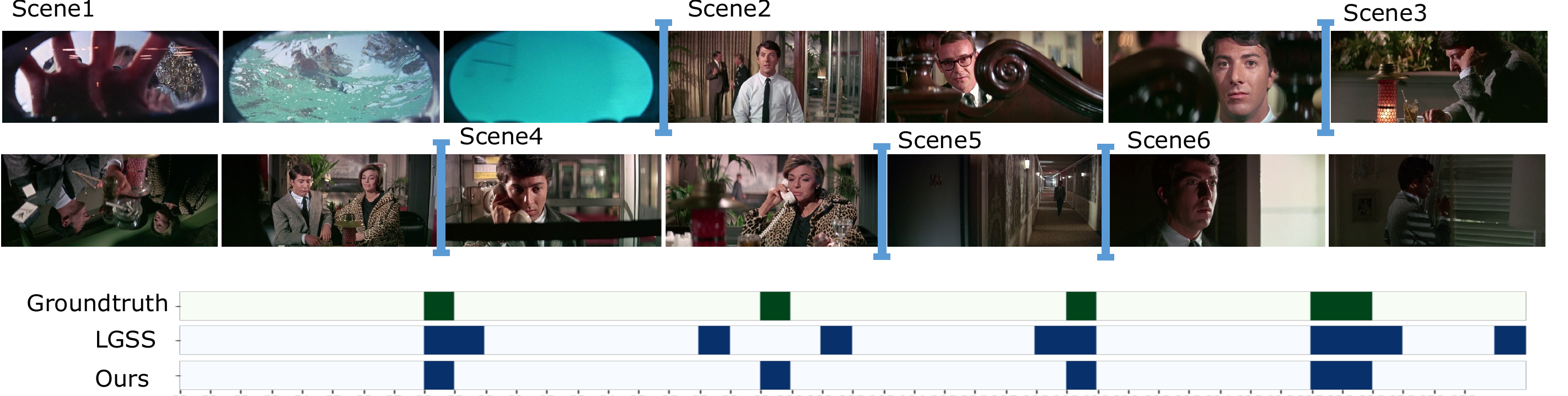}
\label{fig:qualitative_scene}}
\hspace{0.5em}
\subfloat[Qualitative results of event-level boundary detection on Kinetics-GEBD. Compared to the previous supervised SOTA PC, our method achieves sparse boundary detection and predicts boundaries with smaller relative distance to groundtruth.]{\includegraphics[scale=0.59]{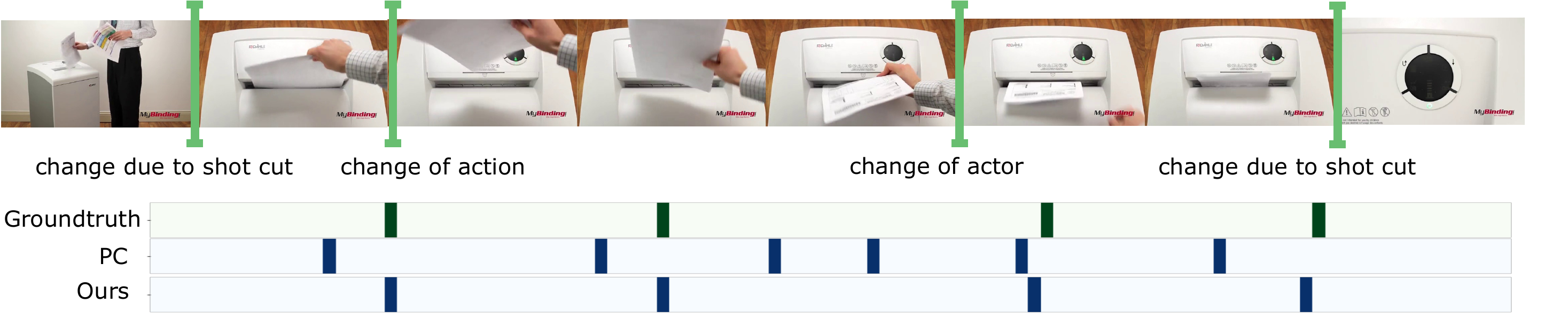}
\label{fig:qualitative_event}}
\caption{Qualitative Evaluation for (a) Scene-level GBD on MovieNet; (b) Event-level GBD on Kinetics-GEBD.}
\label{fig:qualitative}
\end{figure*}

\begin{figure*}[!t]
\centering
\subfloat[{The visualization of coherence scores on MovieNet (scene-level).} We show the raw frames from seven scenes in a movie before the compression with the x-axis showing the shot number (row 1), large redundancy are present in especially scene 6. The groundtruth boundaries of this movie (row 2) further demonstrate the large redundancy in movies. Then, we show the predicted coherence scores for the seven scenes achieves high recall for boundary capture in comparison with the groundtruth (row 3), serving as good prior for feature compression.]{\includegraphics[scale=0.58]{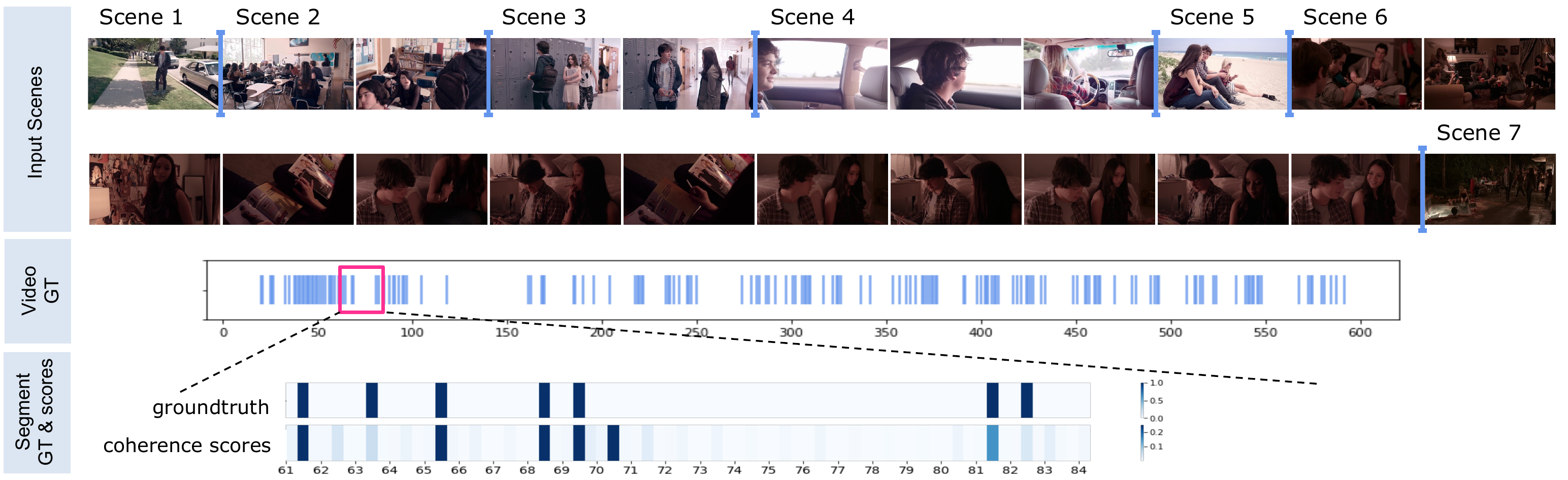}
\label{fig:scene_compression}}
\hspace{0.5em}
\subfloat[The visualization of coherence scores on Kinetics-GEBD (event-level). The video records the event of a woman doing laundry, as shown in raw frames (row 1). The temporal redundancy occurs mainly in clip 3 and 4. Coherence scores (row 3) accurately detects potential boundaries and the redundant clips are allocated into context regions to get clustered into semantics centers. Additionally, boundary regions decided by top-k highest coherence scores are indicated with dark blocks in row 4.]{\includegraphics[scale=0.59]{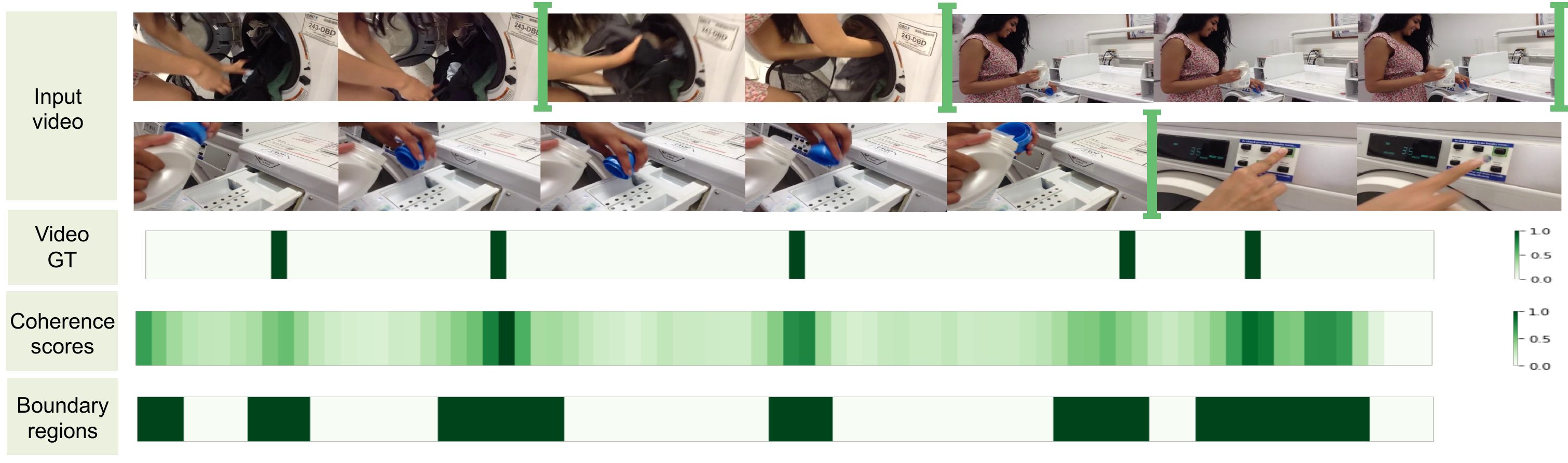}\label{fig:event_compression}}
\caption{The visualization of coherence scores on (a) MovieNet for Scene-level GBD; (b) Kinetics-GEBD for Event-level GBD.}
\label{fig:compression}
\end{figure*}

\textit{Study on input temporal length.}
The input temporal feature length $N$ determines the global temporal receptive field in feature compression, therefore we study the impact of increased temporal field to detection performance in \cref{table:ablation}. The experiments are conducted based on the default choices of N for each benchmark, and we learn that increasing N by $2\times$ or $3\times$ results in severe performance decrease. Decreasing the input length is also harmful to the detection performance. We analyze that boundary detection do not need very large temporal receptive field and increased window size could introduce more noise. 

\begin{figure*}[t]
\begin{center}
\includegraphics[scale=0.58]{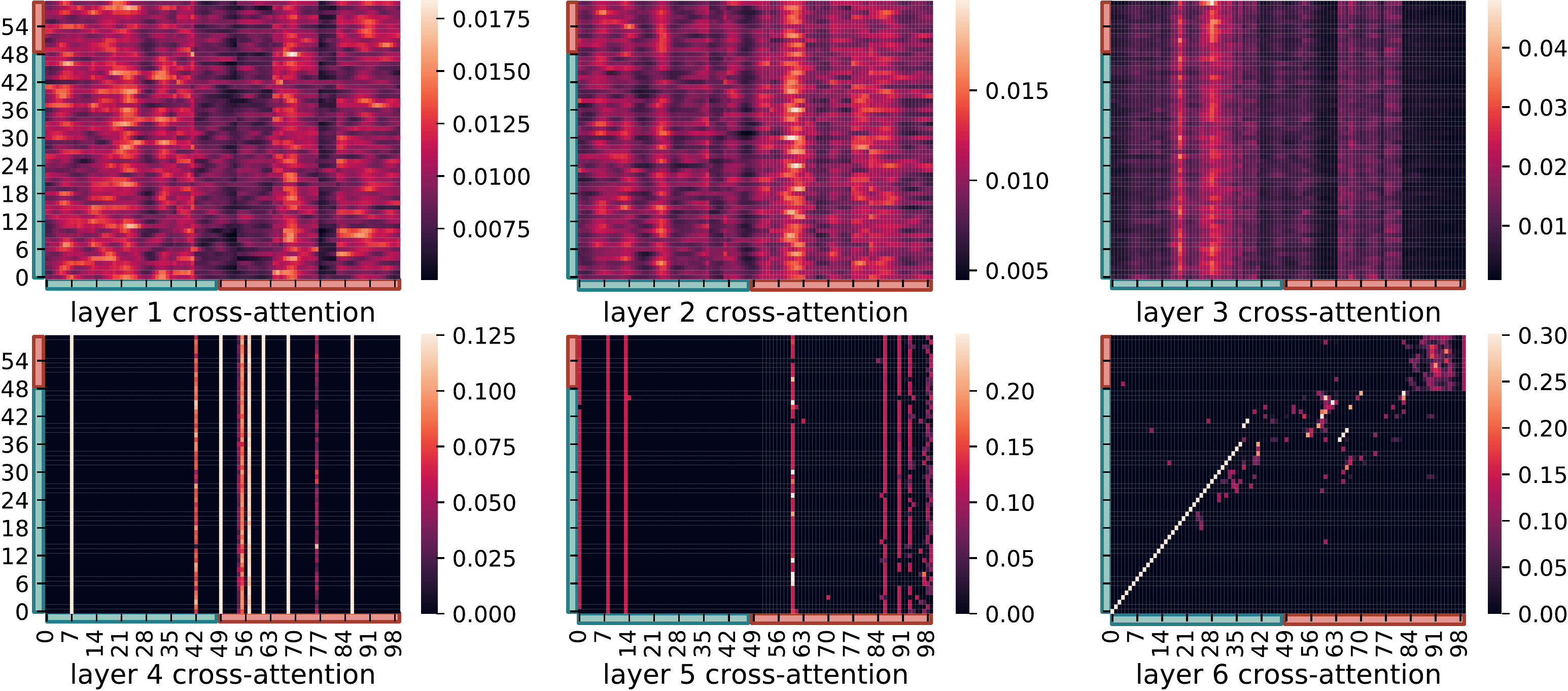}
\end{center}
  \caption{{The Visualization of encoder cross-attention from Layer 1 to Layer 6.} The x-axis indicates the temporal locations in input sequence, and the y-axis indicates the sequence in latent queries. The boundary queries and features are shown with green bars along axes, and the context queries and features are indicated with red bars. The obvious diagonal pattern observed in the last cross-attention shows efficient preservation of boundary features by boundary queries.}
\label{fig:vis_crossattn}
\end{figure*}

\section{Model Visualization}

In this section, we provide qualitative results for event-level and scene-level boundary detection. In addition, comprehensive visualizations for coherence scores and encoder cross-attention are provided to facilitate the understanding of how Temporal Perceiver works. We specifically illustrate how the coherence scores instructs the compression for scene-level and event-level boundary detection. Moreover, we look inside the encoder cross-attention maps for feature compression and show how the alignment loss is enforced on cross-attention activation maps.

\subsection{Qualitative evaluation}
In \cref{fig:qualitative_scene}, we provide the qualitative results on a randomly selected segment from the validation set of MovieNet. We visualize the groundtruth boundaries and the predictions from our model and previous SOTA method LGSS. The LGSS results are reproduced on MovieNet with multi-modal input, whereas our method takes single-stream RGB features only. Compared with LGSS results, our method does not produce duplicates around groundtruth or other within-scene false positives and is able to detect the generic boundary with accurate localization.

To demonstrate the model generalization ability, in \cref{fig:qualitative_event} we also provide the qualitative results on a random video sample from Kinetics-GEBD validation set. Compared to the previous state-of-the-art method PC, the predictions of Temporal Perceiver are more accurate, with less relative distance to groundtruth and handles the relatively intense within-event motion in the third event with fewer false positives.

\subsection{Visualization of coherence scores}
We perform feature compression on boundary and context regions and distinguish these regions based on the coherence estimation scores. \cref{fig:compression} illustrates coherence scores in feature compression process with scene-level and event-level video to demonstrate its effectiveness with different levels of video redundancy. In \cref{fig:scene_compression}, the coherence scores are visualized in row 3. Darker color represents bigger likeliness of frames to cover incoherence regions. In the visualization of video content and groundtruth, we observe larger temporal redundancy for movie clips, especially in scene 6. The coherence scores predict higher likeliness scores near groundtruth boundaries and remain low in scene 6. As a result, the redundant frames in scene 6 are not marked as boundaries and would be clustered into few contextual centers, achieving redundancy compression.

The coherence scores for event-level testing sample is shown in \cref{fig:event_compression}. Compared to scene-level movie clips, although daily event videos are less redundant in temporal aspect, Temporal Perceiver still works for event boundary detection. We observe that similar to feature compression with scene-level sample, the coherence scores (row 3) give coarse yet reliable estimation of boundary regions. We additionally marks the selected boundary regions for event-level sample in row 4. It is observed that all groundtruths are covered in these selected boundary regions. Meanwhile, the temporal redundancy between generic boundaries, such as the repeated back-and-forth movement of pouring detergent in the fourth clip, are excluded and would be clustered into a few centers.

\subsection{Visualization of encoder cross-attention maps}
\cref{fig:vis_crossattn} is the visualization of cross-attention blocks in all 6 layers of encoder on a randomly selected video of Kinetics-GEBD.  In the first 2 layers, the latent queries aggregate information from both the boundary and context regions as an initialization. In the third to fifth layer, latent queries attend to parts of boundary regions and context regions. Beam patterns appear at different parts for each layer without overlap. In the last layer, guided by the diagonal alignment loss, most of the boundary queries aggregate corresponding boundary features in a diagonal-alike pattern, with a few out-of-line attention on neighboring features. The context queries are free of additional supervision and learn to cluster the contextual centers within the context region.

\section{Conclusion}
\label{sec:conclusion}
 In this paper, we have presented Temporal Perceiver, a general architecture for generic boundary detection. It offers an effective pipeline with Transformer architecture and provides a unified framework for the sparse detection of arbitrary generic boundaries. The core contribution is to utilize cross-attention blocks and a small set of latent queries to squeeze redundant video input into a fixed dimension, reducing encoder complexity to linear-level. Moreover, we construct the latent units with boundary and context queries to pattern semantic incoherent boundary features and coherent context features. Furthermore, to facilitate training, we introduce a new alignment loss on encoder cross-attention maps for feature-query alignment. Finally, the sparse detection paradigm of transformer decoder allows our model to be free of post-processing, resulting in a more efficient framework. Experiments show that Temporal Perceiver achieves state-of-the-art results on benchmarks of shot-level, event-level, and scene-level generic boundaries, demonstrating the generalization ability of our model to handle arbitrary generic boundaries.

\ifCLASSOPTIONcompsoc
  \section*{Acknowledgments}
\else
  \section*{Acknowledgment}
\fi

 This work is supported by National Natural Science Foundation of China (No. 62076119, No. 61921006), the Fundamental Research Funds for the Central Universities (No. 020214380091), and Collaborative Innovation Center of Novel Software Technology and Industrialization.

\ifCLASSOPTIONcaptionsoff
  \newpage
\fi

{
\bibliographystyle{IEEEtran}
\bibliography{egbib}

\begin{thebibliography}{10}
\providecommand{\url}[1]{#1}
\csname url@samestyle\endcsname
\providecommand{\newblock}{\relax}
\providecommand{\bibinfo}[2]{#2}
\providecommand{\BIBentrySTDinterwordspacing}{\spaceskip=0pt\relax}
\providecommand{\BIBentryALTinterwordstretchfactor}{4}
\providecommand{\BIBentryALTinterwordspacing}{\spaceskip=\fontdimen2\font plus
\BIBentryALTinterwordstretchfactor\fontdimen3\font minus
  \fontdimen4\font\relax}
\providecommand{\BIBforeignlanguage}[2]{{%
\expandafter\ifx\csname l@#1\endcsname\relax
\typeout{** WARNING: IEEEtran.bst: No hyphenation pattern has been}%
\typeout{** loaded for the language `#1'. Using the pattern for}%
\typeout{** the default language instead.}%
\else
\language=\csname l@#1\endcsname
\fi
#2}}
\providecommand{\BIBdecl}{\relax}
\BIBdecl

\bibitem{DBLP:conf/cvpr/CarreiraZ17}
J.~Carreira and A.~Zisserman, ``Quo vadis, action recognition? {A} new model
  and the kinetics dataset,'' in \emph{{CVPR}}.\hskip 1em plus 0.5em minus
  0.4em\relax {IEEE} Computer Society, 2017, pp. 4724--4733.

\bibitem{DBLP:conf/eccv/WangXW0LTG16}
L.~Wang, Y.~Xiong, Z.~Wang, Y.~Qiao, D.~Lin, X.~Tang, and L.~V. Gool,
  ``Temporal segment networks: Towards good practices for deep action
  recognition,'' in \emph{{ECCV} {(8)}}, ser. Lecture Notes in Computer
  Science, vol. 9912.\hskip 1em plus 0.5em minus 0.4em\relax Springer, 2016,
  pp. 20--36.

\bibitem{DBLP:journals/corr/abs-2203-12602}
Z.~Tong, Y.~Song, J.~Wang, and L.~Wang, ``Videomae: Masked autoencoders are
  data-efficient learners for self-supervised video pre-training,''
  \emph{CoRR}, vol. abs/2203.12602, 2022.

\bibitem{DBLP:conf/iccv/ZhiTWW21}
Y.~Zhi, Z.~Tong, L.~Wang, and G.~Wu, ``Mgsampler: An explainable sampling
  strategy for video action recognition,'' in \emph{{ICCV}}.\hskip 1em plus
  0.5em minus 0.4em\relax {IEEE}, October 2021, pp. 1513--1522.

\bibitem{DBLP:conf/eccv/LiW0W20}
Y.~Li, Z.~Wang, L.~Wang, and G.~Wu, ``Actions as moving points,'' in
  \emph{{ECCV} {(16)}}, ser. Lecture Notes in Computer Science, vol.
  12361.\hskip 1em plus 0.5em minus 0.4em\relax Springer, 2020, pp. 68--84.

\bibitem{DBLP:conf/iccv/TengWLW21}
Y.~Teng, L.~Wang, Z.~Li, and G.~Wu, ``Target adaptive context aggregation for
  video scene graph generation,'' in \emph{{ICCV}}.\hskip 1em plus 0.5em minus
  0.4em\relax {IEEE}, October 2021, pp. 13\,688--13\,697.

\bibitem{DBLP:conf/iccv/JiDN21}
J.~Ji, R.~Desai, and J.~C. Niebles, ``Detecting human-object relationships in
  videos,'' in \emph{{ICCV}}.\hskip 1em plus 0.5em minus 0.4em\relax {IEEE},
  2021, pp. 8086--8096.

\bibitem{DBLP:conf/cvpr/RaiCJDKI0N21}
N.~Rai, H.~Chen, J.~Ji, R.~Desai, K.~Kozuka, S.~Ishizaka, E.~Adeli, and J.~C.
  Niebles, ``Home action genome: Cooperative compositional action
  understanding,'' in \emph{{CVPR}}.\hskip 1em plus 0.5em minus 0.4em\relax
  Computer Vision Foundation / {IEEE}, 2021, pp. 11\,184--11\,193.

\bibitem{DBLP:conf/cvpr/JiK0N20}
J.~Ji, R.~Krishna, L.~Fei{-}Fei, and J.~C. Niebles, ``Action genome: Actions as
  compositions of spatio-temporal scene graphs,'' in \emph{{CVPR}}.\hskip 1em
  plus 0.5em minus 0.4em\relax Computer Vision Foundation / {IEEE}, 2020, pp.
  10\,233--10\,244.

\bibitem{DBLP:conf/nips/SimonyanZ14}
K.~Simonyan and A.~Zisserman, ``Two-stream convolutional networks for action
  recognition in videos,'' in \emph{{NIPS}}, 2014, pp. 568--576.

\bibitem{DBLP:conf/cvpr/WangL0G18}
L.~Wang, W.~Li, W.~Li, and L.~V. Gool, ``Appearance-and-relation networks for
  video classification,'' in \emph{{CVPR}}.\hskip 1em plus 0.5em minus
  0.4em\relax Computer Vision Foundation / {IEEE} Computer Society, 2018, pp.
  1430--1439.

\bibitem{DBLP:conf/iccv/Feichtenhofer0M19}
C.~Feichtenhofer, H.~Fan, J.~Malik, and K.~He, ``Slowfast networks for video
  recognition,'' in \emph{{ICCV}}.\hskip 1em plus 0.5em minus 0.4em\relax
  {IEEE}, 2019, pp. 6201--6210.

\bibitem{DBLP:conf/iccv/ZhiT0W21}
Y.~Zhi, Z.~Tong, L.~Wang, and G.~Wu, ``Mgsampler: An explainable sampling
  strategy for video action recognition,'' in \emph{{ICCV}}.\hskip 1em plus
  0.5em minus 0.4em\relax {IEEE}, 2021, pp. 1493--1502.

\bibitem{DBLP:conf/cvpr/LiJSZKW20}
Y.~Li, B.~Ji, X.~Shi, J.~Zhang, B.~Kang, and L.~Wang, ``{TEA:} temporal
  excitation and aggregation for action recognition,'' in \emph{{CVPR}}.\hskip
  1em plus 0.5em minus 0.4em\relax Computer Vision Foundation / {IEEE}, 2020,
  pp. 906--915.

\bibitem{DBLP:conf/cvpr/0002TJW21}
L.~Wang, Z.~Tong, B.~Ji, and G.~Wu, ``{TDN:} temporal difference networks for
  efficient action recognition,'' in \emph{{CVPR}}.\hskip 1em plus 0.5em minus
  0.4em\relax Computer Vision Foundation / {IEEE}, 2021, pp. 1895--1904.

\bibitem{DBLP:conf/cvpr/ChaoVSRDS18}
Y.~Chao, S.~Vijayanarasimhan, B.~Seybold, D.~A. Ross, J.~Deng, and
  R.~Sukthankar, ``Rethinking the faster {R-CNN} architecture for temporal
  action localization,'' in \emph{{CVPR}}.\hskip 1em plus 0.5em minus
  0.4em\relax Computer Vision Foundation / {IEEE} Computer Society, 2018, pp.
  1130--1139.

\bibitem{DBLP:conf/cvpr/Lin0LWTWLHF21}
C.~Lin, C.~Xu, D.~Luo, Y.~Wang, Y.~Tai, C.~Wang, J.~Li, F.~Huang, and Y.~Fu,
  ``Learning salient boundary feature for anchor-free temporal action
  localization,'' in \emph{{CVPR}}.\hskip 1em plus 0.5em minus 0.4em\relax
  Computer Vision Foundation / {IEEE}, 2021, pp. 3320--3329.

\bibitem{DBLP:conf/iccv/TanTWW21}
J.~Tan, J.~Tang, L.~Wang, and G.~Wu, ``Relaxed transformer decoders for direct
  action proposal generation,'' in \emph{{ICCV}}, 2021, pp. 13\,526--13\,535.

\bibitem{Zhao_2022_CVPR}
J.~Zhao, Y.~Zhang, X.~Li, H.~Chen, B.~Shuai, M.~Xu, C.~Liu, K.~Kundu, Y.~Xiong,
  D.~Modolo, I.~Marsic, C.~G.~M. Snoek, and J.~Tighe, ``Tuber: Tubelet
  transformer for video action detection,'' in \emph{{CVPR}}.\hskip 1em plus
  0.5em minus 0.4em\relax Computer Vision Foundation / {IEEE}, June 2022, pp.
  13\,598--13\,607.

\bibitem{DBLP:conf/iccv/KalogeitonWFS17a}
V.~Kalogeiton, P.~Weinzaepfel, V.~Ferrari, and C.~Schmid, ``Action tubelet
  detector for spatio-temporal action localization,'' in \emph{{ICCV}}.\hskip
  1em plus 0.5em minus 0.4em\relax {IEEE} Computer Society, 2017, pp.
  4415--4423.

\bibitem{DBLP:conf/iccv/SinghSSTC17}
G.~Singh, S.~Saha, M.~Sapienza, P.~H.~S. Torr, and F.~Cuzzolin, ``Online
  real-time multiple spatiotemporal action localisation and prediction,'' in
  \emph{{ICCV}}.\hskip 1em plus 0.5em minus 0.4em\relax {IEEE} Computer
  Society, 2017, pp. 3657--3666.

\bibitem{DBLP:conf/eccv/Gabeur0AS20}
V.~Gabeur, C.~Sun, K.~Alahari, and C.~Schmid, ``Multi-modal transformer for
  video retrieval,'' in \emph{{ECCV} {(4)}}, ser. Lecture Notes in Computer
  Science, vol. 12349.\hskip 1em plus 0.5em minus 0.4em\relax Springer, 2020,
  pp. 214--229.

\bibitem{DBLP:conf/cvpr/LeiLZGBB021}
J.~Lei, L.~Li, L.~Zhou, Z.~Gan, T.~L. Berg, M.~Bansal, and J.~Liu, ``Less is
  more: Clipbert for video-and-language learning via sparse sampling,'' in
  \emph{{CVPR}}.\hskip 1em plus 0.5em minus 0.4em\relax Computer Vision
  Foundation / {IEEE}, 2021, pp. 7331--7341.

\bibitem{Ge_2022_CVPR}
Y.~Ge, Y.~Ge, X.~Liu, D.~Li, Y.~Shan, X.~Qie, and P.~Luo, ``Bridging video-text
  retrieval with multiple choice questions,'' in \emph{{CVPR}}.\hskip 1em plus
  0.5em minus 0.4em\relax Computer Vision Foundation / {IEEE}, June 2022, pp.
  16\,167--16\,176.

\bibitem{DBLP:conf/cvpr/ZhangDWWD19}
D.~Zhang, X.~Dai, X.~Wang, Y.~Wang, and L.~S. Davis, ``{MAN:} moment alignment
  network for natural language moment retrieval via iterative graph
  adjustment,'' in \emph{{CVPR}}.\hskip 1em plus 0.5em minus 0.4em\relax
  Computer Vision Foundation / {IEEE}, 2019, pp. 1247--1257.

\bibitem{DBLP:conf/aaai/ZhangPFL20}
S.~Zhang, H.~Peng, J.~Fu, and J.~Luo, ``Learning 2d temporal adjacent networks
  for moment localization with natural language,'' in \emph{{AAAI}}.\hskip 1em
  plus 0.5em minus 0.4em\relax {AAAI} Press, 2020, pp. 12\,870--12\,877.

\bibitem{DBLP:conf/emnlp/XiaoCSZ021}
S.~Xiao, L.~Chen, J.~Shao, Y.~Zhuang, and J.~Xiao, ``Natural language video
  localization with learnable moment proposals,'' in \emph{{EMNLP}
  {(1)}}.\hskip 1em plus 0.5em minus 0.4em\relax Association for Computational
  Linguistics, 2021, pp. 4008--4017.

\bibitem{DBLP:conf/aaai/00010WLW22}
Z.~Wang, L.~Wang, T.~Wu, T.~Li, and G.~Wu, ``Negative sample matters: {A}
  renaissance of metric learning for temporal grounding,'' in
  \emph{{AAAI}}.\hskip 1em plus 0.5em minus 0.4em\relax {AAAI} Press, 2022, pp.
  2613--2623.

\bibitem{yang2022tubedetr}
A.~Yang, A.~Miech, J.~Sivic, I.~Laptev, and C.~Schmid, ``Tubedetr:
  Spatio-temporal video grounding with transformers,'' in \emph{{CVPR}}.\hskip
  1em plus 0.5em minus 0.4em\relax Computer Vision Foundation / {IEEE}, 2022,
  pp. 16\,442--16\,453.

\bibitem{DBLP:conf/iccv/ShouLWGF21}
M.~Z. Shou, S.~W. Lei, W.~Wang, D.~Ghadiyaram, and M.~Feiszli, ``Generic event
  boundary detection: A benchmark for event segmentation,'' in \emph{{ICCV}},
  2021, pp. 8075--8084.

\bibitem{DBLP:conf/iccv/LinLLDW19}
T.~Lin, X.~Liu, X.~Li, E.~Ding, and S.~Wen, ``{BMN:} boundary-matching network
  for temporal action proposal generation,'' in \emph{{ICCV}}, 2019, pp.
  3888--3897.

\bibitem{DBLP:conf/aaai/GaoSWLYGZ20}
J.~Gao, Z.~Shi, G.~Wang, J.~Li, Y.~Yuan, S.~Ge, and X.~Zhou, ``Accurate
  temporal action proposal generation with relation-aware pyramid network,'' in
  \emph{{AAAI}}.\hskip 1em plus 0.5em minus 0.4em\relax {AAAI} Press, 2020, pp.
  10\,810--10\,817.

\bibitem{DBLP:conf/eccv/WangGWLW20}
Z.~Wang, Z.~Gao, L.~Wang, Z.~Li, and G.~Wu, ``Boundary-aware cascade networks
  for temporal action segmentation,'' in \emph{{ECCV}}, 2020, pp. 34--51.

\bibitem{DBLP:conf/eccv/BaiWTYLL20}
Y.~Bai, Y.~Wang, Y.~Tong, Y.~Yang, Q.~Liu, and J.~Liu, ``Boundary content graph
  neural network for temporal action proposal generation,'' in \emph{{ECCV}
  {(28)}}, ser. Lecture Notes in Computer Science, vol. 12373.\hskip 1em plus
  0.5em minus 0.4em\relax Springer, 2020, pp. 121--137.

\bibitem{DBLP:conf/cvpr/XuZRTG20}
M.~Xu, C.~Zhao, D.~S. Rojas, A.~K. Thabet, and B.~Ghanem, ``{G-TAD:} sub-graph
  localization for temporal action detection,'' in \emph{{CVPR}}.\hskip 1em
  plus 0.5em minus 0.4em\relax Computer Vision Foundation / {IEEE}, 2020, pp.
  10\,153--10\,162.

\bibitem{DBLP:conf/cvpr/DeliegeCGSDNGMD21}
A.~Deli{\`{e}}ge, A.~Cioppa, S.~Giancola, M.~J. Seikavandi, J.~V. Dueholm,
  K.~Nasrollahi, B.~Ghanem, T.~B. Moeslund, and M.~V. Droogenbroeck,
  ``Soccernet-v2: {A} dataset and benchmarks for holistic understanding of
  broadcast soccer videos,'' in \emph{{CVPR} Workshops}.\hskip 1em plus 0.5em
  minus 0.4em\relax Computer Vision Foundation / {IEEE}, 2021, pp. 4508--4519.

\bibitem{DBLP:conf/eccv/HuangXRWL20}
Q.~Huang, Y.~Xiong, A.~Rao, J.~Wang, and D.~Lin, ``Movienet: {A} holistic
  dataset for movie understanding,'' in \emph{{ECCV}}, 2020, pp. 709--727.

\bibitem{scikit}
S.-V. Developers, ``Scikit-video: Video processing in python,,''
  \url{https://github.com/scikit-video/scikit-video.}, 2015.

\bibitem{DBLP:conf/cvpr/RaoXXXHZL20}
A.~Rao, L.~Xu, Y.~Xiong, G.~Xu, Q.~Huang, B.~Zhou, and D.~Lin, ``A
  local-to-global approach to multi-modal movie scene segmentation,'' in
  \emph{{CVPR}}, 2020, pp. 10\,143--10\,152.

\bibitem{DBLP:conf/cvpr/ChenNFZBH21}
S.~Chen, X.~Nie, D.~Fan, D.~Zhang, V.~Bhat, and R.~Hamid, ``Shot contrastive
  self-supervised learning for scene boundary detection,'' in \emph{{CVPR}},
  2021, pp. 9796--9805.

\bibitem{DBLP:conf/cvpr/ShaoZDL20}
D.~Shao, Y.~Zhao, B.~Dai, and D.~Lin, ``Intra- and inter-action understanding
  via temporal action parsing,'' in \emph{{CVPR}}, 2020, pp. 727--736.

\bibitem{DBLP:conf/cvpr/HeZRS16}
K.~He, X.~Zhang, S.~Ren, and J.~Sun, ``Deep residual learning for image
  recognition,'' in \emph{{CVPR}}.\hskip 1em plus 0.5em minus 0.4em\relax
  {IEEE} Computer Society, 2016, pp. 770--778.

\bibitem{DBLP:conf/cvpr/GhadiyaramTM19}
D.~Ghadiyaram, D.~Tran, and D.~Mahajan, ``Large-scale weakly-supervised
  pre-training for video action recognition,'' in \emph{{CVPR}}.\hskip 1em plus
  0.5em minus 0.4em\relax Computer Vision Foundation / {IEEE}, 2019, pp.
  12\,046--12\,055.

\bibitem{DBLP:conf/iccv/TranWFT19}
D.~Tran, H.~Wang, M.~Feiszli, and L.~Torresani, ``Video classification with
  channel-separated convolutional networks,'' in \emph{{ICCV}}.\hskip 1em plus
  0.5em minus 0.4em\relax {IEEE}, 2019, pp. 5551--5560.

\bibitem{PyScenedetect}
B.~Castellano, ``Pyscenedetect: Video scene cut detection and analysis tool,''
  \url{https://github.com/Breakthrough/PySceneDetect}, 2014.

\bibitem{DBLP:conf/eccv/LeaRVH16}
C.~Lea, A.~Reiter, R.~Vidal, and G.~D. Hager, ``Segmental spatiotemporal cnns
  for fine-grained action segmentation,'' in \emph{{ECCV} {(3)}}, ser. Lecture
  Notes in Computer Science, vol. 9907.\hskip 1em plus 0.5em minus 0.4em\relax
  Springer, 2016, pp. 36--52.

\bibitem{DBLP:conf/eccv/LinZSWY18}
T.~Lin, X.~Zhao, H.~Su, C.~Wang, and M.~Yang, ``{BSN:} boundary sensitive
  network for temporal action proposal generation,'' in \emph{{ECCV} {(4)}},
  ser. Lecture Notes in Computer Science, vol. 11208.\hskip 1em plus 0.5em
  minus 0.4em\relax Springer, 2018, pp. 3--21.

\bibitem{DBLP:conf/cvpr/DingX18}
L.~Ding and C.~Xu, ``Weakly-supervised action segmentation with iterative soft
  boundary assignment,'' in \emph{{CVPR}}.\hskip 1em plus 0.5em minus
  0.4em\relax Computer Vision Foundation / {IEEE} Computer Society, 2018, pp.
  6508--6516.

\bibitem{DBLP:conf/eccv/HuangFN16}
D.~Huang, L.~Fei{-}Fei, and J.~C. Niebles, ``Connectionist temporal modeling
  for weakly supervised action labeling,'' in \emph{{ECCV}}, 2016, pp.
  137--153.

\bibitem{DBLP:conf/ism/RotmanPA16}
D.~Rotman, D.~Porat, and G.~Ashour, ``Robust and efficient video scene
  detection using optimal sequential grouping,'' in \emph{{ISM}}, 2016, pp.
  275--280.

\bibitem{DBLP:conf/mm/BaraldiGC15}
L.~Baraldi, C.~Grana, and R.~Cucchiara, ``A deep siamese network for scene
  detection in broadcast videos,'' in \emph{{ACM} Multimedia}, 2015, pp.
  1199--1202.

\bibitem{DBLP:journals/tcsv/LiuKBP21}
D.~Liu, N.~Kamath, S.~Bhattacharya, and R.~Puri, ``Adaptive context reading
  network for movie scene detection,'' \emph{{IEEE} Trans. Circuits Syst. Video
  Technol.}, vol.~31, no.~9, pp. 3559--3574, 2021.

\bibitem{DBLP:conf/icmcs/RuiHM98}
Y.~Rui, T.~S. Huang, and S.~Mehrotra, ``Exploring video structure beyond the
  shots,'' in \emph{{ICMCS}}.\hskip 1em plus 0.5em minus 0.4em\relax {IEEE}
  Computer Society, 1998, pp. 237--240.

\bibitem{DBLP:conf/cvpr/RasheedS03}
Z.~Rasheed and M.~Shah, ``Scene detection in hollywood movies and {TV} shows,''
  in \emph{{CVPR} {(2)}}.\hskip 1em plus 0.5em minus 0.4em\relax {IEEE}
  Computer Society, 2003, pp. 343--350.

\bibitem{DBLP:journals/tmm/ChasanisLG09}
V.~Chasanis, A.~Likas, and N.~P. Galatsanos, ``Scene detection in videos using
  shot clustering and sequence alignment,'' \emph{{IEEE} Trans. Multim.},
  vol.~11, no.~1, pp. 89--100, 2009.

\bibitem{DBLP:conf/icmcs/HanW11}
B.~Han and W.~Wu, ``Video scene segmentation using a novel boundary evaluation
  criterion and dynamic programming,'' in \emph{{ICME}}, 2011, pp. 1--6.

\bibitem{DBLP:conf/cvpr/TapaswiBS14}
M.~Tapaswi, M.~B{\"{a}}uml, and R.~Stiefelhagen, ``Storygraphs: Visualizing
  character interactions as a timeline,'' in \emph{{CVPR}}, 2014, pp. 827--834.

\bibitem{DBLP:conf/nips/VaswaniSPUJGKP17}
A.~Vaswani, N.~Shazeer, N.~Parmar, J.~Uszkoreit, L.~Jones, A.~N. Gomez,
  L.~Kaiser, and I.~Polosukhin, ``Attention is all you need,'' in
  \emph{{NIPS}}, 2017, pp. 5998--6008.

\bibitem{DBLP:conf/naacl/DevlinCLT19}
J.~Devlin, M.~Chang, K.~Lee, and K.~Toutanova, ``{BERT:} pre-training of deep
  bidirectional transformers for language understanding,'' in
  \emph{{NAACL-HLT}}, 2019, pp. 4171--4186.

\bibitem{DBLP:conf/eccv/CarionMSUKZ20}
N.~Carion, F.~Massa, G.~Synnaeve, N.~Usunier, A.~Kirillov, and S.~Zagoruyko,
  ``End-to-end object detection with transformers,'' in \emph{{ECCV} {(1)}},
  ser. Lecture Notes in Computer Science, vol. 12346.\hskip 1em plus 0.5em
  minus 0.4em\relax Springer, 2020, pp. 213--229.

\bibitem{DBLP:conf/iclr/DosovitskiyB0WZ21}
A.~Dosovitskiy, L.~Beyer, A.~Kolesnikov, D.~Weissenborn, X.~Zhai,
  T.~Unterthiner, M.~Dehghani, M.~Minderer, G.~Heigold, S.~Gelly, J.~Uszkoreit,
  and N.~Houlsby, ``An image is worth 16x16 words: Transformers for image
  recognition at scale,'' in \emph{{ICLR}}, 2021.

\bibitem{DBLP:conf/iccv/LiuLCHWZLG21}
Z.~Liu, Y.~Lin, Y.~Cao, H.~Hu, Y.~Wei, Z.~Zhang, S.~Lin, and B.~Guo, ``Swin
  transformer: Hierarchical vision transformer using shifted windows,'' in
  \emph{{ICCV}}, 2021, pp. 10\,012--10\,022.

\bibitem{DBLP:journals/corr/abs-2106-13230}
Z.~Liu, J.~Ning, Y.~Cao, Y.~Wei, Z.~Zhang, S.~Lin, and H.~Hu, ``Video swin
  transformer,'' \emph{CoRR}, vol. abs/2106.13230, 2021.

\bibitem{DBLP:conf/icml/TouvronCDMSJ21}
H.~Touvron, M.~Cord, M.~Douze, F.~Massa, A.~Sablayrolles, and H.~J{\'{e}}gou,
  ``Training data-efficient image transformers {\&} distillation through
  attention,'' in \emph{{ICML}}, ser. Proceedings of Machine Learning Research,
  vol. 139.\hskip 1em plus 0.5em minus 0.4em\relax {PMLR}, 2021, pp.
  10\,347--10\,357.

\bibitem{DBLP:conf/iccv/ArnabDHSLS21}
A.~Arnab, M.~Dehghani, G.~Heigold, C.~Sun, M.~Lu\v{c}i\'c, and C.~Schmid,
  ``Vivit: {A} video vision transformer,'' in \emph{{ICCV}}.\hskip 1em plus
  0.5em minus 0.4em\relax {IEEE}, October 2021, pp. 6836--6846.

\bibitem{DBLP:conf/iclr/ZhuSLLWD21}
X.~Zhu, W.~Su, L.~Lu, B.~Li, X.~Wang, and J.~Dai, ``Deformable {DETR:}
  deformable transformers for end-to-end object detection,'' in \emph{{ICLR}},
  2021.

\bibitem{DBLP:conf/iccv/WangYCFY21}
T.~Wang, L.~Yuan, Y.~Chen, J.~Feng, and S.~Yan, ``Pnp-detr: Towards efficient
  visual analysis with transformers,'' in \emph{{ICCV}}.\hskip 1em plus 0.5em
  minus 0.4em\relax {IEEE}, October 2021, pp. 4661--4670.

\bibitem{DBLP:conf/icml/JaegleGBVZC21}
A.~Jaegle, F.~Gimeno, A.~Brock, O.~Vinyals, A.~Zisserman, and J.~Carreira,
  ``Perceiver: General perception with iterative attention,'' in \emph{{ICML}},
  2021, pp. 4651--4664.

\bibitem{DBLP:journals/corr/abs-2107-14795}
A.~Jaegle, S.~Borgeaud, J.~Alayrac, C.~Doersch, C.~Ionescu, D.~Ding,
  S.~Koppula, D.~Zoran, A.~Brock, E.~Shelhamer, O.~J. H{\'{e}}naff, M.~M.
  Botvinick, A.~Zisserman, O.~Vinyals, and J.~Carreira, ``Perceiver {IO:} {A}
  general architecture for structured inputs {\&} outputs,'' \emph{CoRR}, vol.
  abs/2107.14795, 2021.

\bibitem{DBLP:conf/cvpr/DengDSLL009}
J.~Deng, W.~Dong, R.~Socher, L.~Li, K.~Li, and L.~Fei{-}Fei, ``Imagenet: {A}
  large-scale hierarchical image database,'' in \emph{{CVPR}}.\hskip 1em plus
  0.5em minus 0.4em\relax {IEEE} Computer Society, 2009, pp. 248--255.

\bibitem{DBLP:journals/corr/KayCSZHVVGBNSZ17}
W.~Kay, J.~Carreira, K.~Simonyan, B.~Zhang, C.~Hillier, S.~Vijayanarasimhan,
  F.~Viola, T.~Green, T.~Back, P.~Natsev, M.~Suleyman, and A.~Zisserman, ``The
  kinetics human action video dataset,'' \emph{CoRR}, vol. abs/1705.06950,
  2017.

\bibitem{DBLP:conf/cvpr/CioppaDGGDGM20}
A.~Cioppa, A.~Deli{\`{e}}ge, S.~Giancola, B.~Ghanem, M.~V. Droogenbroeck,
  R.~Gade, and T.~B. Moeslund, ``A context-aware loss function for action
  spotting in soccer videos,'' in \emph{{CVPR}}.\hskip 1em plus 0.5em minus
  0.4em\relax Computer Vision Foundation / {IEEE}, 2020, pp. 13\,123--13\,133.

\bibitem{DBLP:journals/corr/abs-2206-12634}
D.~Hong, X.~Ma, X.~Wang, C.~Li, Y.~Wang, and L.~Wen, ``Sc-transformer++:
  Structured context transformer for generic event boundary detection,''
  \emph{CoRR}, vol. abs/2206.12634, 2022.

\end{thebibliography}
}

\end{document}